\documentclass[review]{elsarticle}

\usepackage{lineno,hyperref}
\modulolinenumbers[5]

\journal{Pattern Recognition}

\usepackage[normalem]{ulem}

\usepackage{amsfonts}
\usepackage{amsmath}
\usepackage{xcolor}
\usepackage{enumitem}
\usepackage[font=footnotesize]{caption}
\usepackage{float}
\usepackage{bm}

\newcommand{\fig}[1]{Fig.~\ref{#1}}
\newcommand{\eq}[1]{Eq.~(\ref{#1})}

\newcommand{\sect}[1]{Section~\ref{#1}}

\newcommand{\figw}{0.5\columnwidth}

\newcommand{\ben}{\begin{enumerate}}
	\newcommand{\een}{\end{enumerate}}
\newcommand{\be}{\begin{equation}}
	\newcommand{\ee}{\end{equation}}
\newcommand{\bea}{\begin{eqnarray}}
	\newcommand{\eea}{\end{eqnarray}}
\newcommand{\bc}{\begin{cases}}
	\newcommand{\ec}{\end{cases}}
\newcommand{\bi}{\begin{itemize}}
	\newcommand{\ei}{\end{itemize}}

\newcommand{\cutoffone}{f_{\rm c1}}
\newcommand{\cutofftwo}{f_{\rm c2}}
\newcommand{\ncycles}{N_{\rm c}}
\newcommand{\ntest}{N_{\rm t}}
\newcommand{\nsamples}{N_{\rm s}}
\newcommand{\nkernels}[1]{N_{{\rm k}#1}}
\newcommand{\mycorr}{\varphi}
\newcommand{\SVMfeatmap}{\Phi}
\newcommand{\SVMkernel}{\Psi}
\newcommand{\PCA}{\Upsilon}
\newcommand{\decisionfunc}{d}
\newcommand{\scorefunc}{h}
\newcommand{\positiveC}{C_1}
\newcommand{\negativeC}{C_0}

\newcommand{\fraName}{\mbox{IDNet}}

\newcommand{\bre}{\begin{bf}\begin{color}{blue}}
		\newcommand{\ere}{\end{color} \end{bf}}

\renewcommand{\pmb}{\bm}

\begin{document}

\begin{frontmatter}
	
	\title{\fraName: Smartphone-based Gait Recognition \\ with Convolutional Neural Networks}
	
	\author{Matteo~Gadaleta\corref{mycorrespondingauthor}}
	\cortext[mycorrespondingauthor]{Corresponding author}
	\ead{gadaleta@dei.unipd.it}
	
	\author{Michele~Rossi\corref{}}
	\ead{rossi@dei.unipd.it}
	
	\address{ Dept. of Information Engineering, University of Padova, via Gradenigo 6/b, 35131, Padova, Italy }


	\begin{abstract}
Here, we present \fraName, a user authentication framework from smartphone-acquired motion signals. Its goal is to recognize a target user from their way of walking, using the accelerometer and gyroscope (inertial) signals provided by a commercial smartphone worn in the front pocket of the user's trousers. \fraName\ features several innovations including: {\it i)} a robust and \mbox{smartphone-orientation-independent} walking cycle extraction block, {\it ii)} a novel feature extractor based on convolutional neural networks, {\it iii)} a \mbox{one-class} support vector machine to classify walking cycles, and the coherent integration of these into {\it iv)} a \mbox{multi-stage} authentication technique. \fraName\ is the first system that exploits a deep learning approach as universal feature extractors for gait recognition, and that combines classification results from subsequent walking cycles into a \mbox{multi-stage} decision making framework. Experimental results show the superiority of our approach against \mbox{state-of-the-art} techniques, leading to misclassification rates (either false negatives or positives) smaller than $0.15$\% with fewer than five walking cycles. Design choices are discussed and motivated throughout, assessing their impact on the user authentication performance.

	\end{abstract}
	
	
\end{frontmatter}





\section{Introduction}
\label{sec:introduction}

Wearable technology is advancing at a very fast pace. Many wearable devices, such as smart watches and wristbands are currently available in the consumer market and they often possess miniaturized inertial motion sensors (accelerometer and gyroscope) as well as other sensing hardware capable of gathering  biological signs such as photoplethysmographic signals, skin temperature and so forth. Other wearables, such as commercial physiological monitors, deliver a number of vitals via their wireless interfaces, including electrocardiogram, heart rate, chest motion, etc. The same holds true for recent smartphones, that allow for the collection of user's feedback and for the realtime assessment of their health condition. They also feature sophisticated sensing technology, among which we consider inertial sensors. 
%
With sensing technology growing at a fast pace, two major problems are related to the analysis of wearable data and to the authentication of the mobile users who provide it, so that we can assess with reasonably high accuracy whether the data sources are genuine. Notably, certifying the data sources is a necessary step toward the widespread use of this technology in the medical field and, in this paper, we develop the user authentication technology that is required to make this possible.
A great deal of work has been carried out on gait recognition in the last decade~\cite{sprager2015inertial}. In general, biometric gait recognition can be grouped into three main categories: 1) computer vision based, 2) floor sensor based and 3) wearable sensor based~\cite{gafurov2007survey}. Most of the recent work belongs to the first category, where image and video analysis are performed to infer the user identity~\cite{zeng2014silhouette,luo2016robust,xing2016complete,chen2016uncooperative,choudhury2015robust}. Nevertheless, user authentication from wearables is a sensible approach in those scenarios where the deployment of cameras in not possible.

In this paper, we propose \fraName\ (IDentification Network), a new system for the authentication of mobile users from  \mbox{smartphone-acquired} motion data. As shown in~\cite{whittle2007analysis,chan2012evaluating}, modern phones possess highly accurate inertial sensors, which allow for \mbox{non-obtrusive} gait biometrics. \fraName\ leverages deep Convolutional Neural Networks (CNN)~\cite{razavian2014computer} and tools from machine learning, such as Support Vector Machines (SVM) \cite{Bishop-2007}, combining them in an innovative fashion. Specifically, we develop algorithms for 1) walking cycle extraction, 2) feature identification and, finally, 3) user authentication. CNNs are used as {\it universal} feature extractors to discriminate gait signatures from different subjects. Single- as well as \mbox{multi-stage} classifiers are finally combined with CNNs to authenticate the user through the accumulation of scores from subsequent walking cycles. As shown in \sect{sec:multi-class-classification}, our solution authenticates the target user with high accuracy and outperforms state-of-the-art techniques such as~\cite{thang2012gait, nickel2012authentication, watanabe2014influence, choi2014biometric, ren2013smartphone, Sprager-15}. 
The main contributions of this paper are:
\bi
\item The design and validation of an original preprocessing techniques that includes: a robust algorithm for the extraction of walking cycles and an original transformation to move smartphone acquired motion signals into an orientation invariant reference system. Subsequent processing is carried out within this reference system, as this considerably improves authentication results, see \sect{sec:signal_acquisition_and_processing}. As opposed to making motion data orientation independent, previous papers either use data acquired from a sensor in a known and fixed position~\cite{choi2014biometric,ren2013smartphone,chan2013smart,thang2012gait,nickel2012authentication,huang2012gait,nickel2011scenario,nickel2011using}, or use orientation independent features at the cost of losing information about the direction of the forces~\cite{kobayashi2011rotation}.
\item The design of a new CNN-based feature extraction tool, which is trained only once on a representative set of users and then used at runtime as a {\it universal} feature extractor, see \sect{sec:multi-class-classification}. Note that with CNNs, statistical features are automatically extracted as part of the CNN training phase (automatic feature engineering) as opposed to the selection of predefined and often arbitrary features, as commonly done in the literature \cite{watanabe2014influence,choi2014biometric,chan2013smart,nickel2012authentication}.
\item The combination of CNN-extracted features with a \mbox{one-class} SVM classifier~\cite{scholkopf2001SVM}, which is solely trained on the target subject, see \sect{sec:one_class_classification}. The resulting SVM scores are then accumulated across multiple walking cycles to get higher accuracies, through a new multi-stage identification framework, see \sect{sec:sequential_analysis}.
\item The coherent integration of these techniques into the \fraName\ authentication framework, that uses smartphone-acquired accelerometer and gyroscope motion data. We also show that the integration of gyroscope data provides further performance improvements, see \sect{sec:network_optimization}.
\item The experimental validation of \fraName, proving its superiority against solutions from the \mbox{state-of-the-art}, see \sect{sec:multi-class-classification}, and achieving authentication errors below $0.15$\% using fewer than five walking cycles, see \sect{sec:sequential_analysis}.
\ei




\section{Related Work}
\label{sec:related_work}

Interest in gait analysis began in the 60's, when walking patterns from healthy people, termed as normal patterns, were investigated by Murray et al.~\cite{murray1964walking}. These measurements were performed through the analysis of photos acquired using interrupted light photography. Murray compared normal gait parameters against those from pathologic gaits~\cite{murray1967gait} and showed that gait is unique to each individual. Since then, human identification through gait recognition has been enjoying a growing interest. 
Most recent works are based on computer vision~\cite{nixon2006human,gafurov2007survey}.
Currently, \mbox{multi-view} gait recognition problem and condition invariance (e.g., clothing or carried items, walking speed, view angle, etc.) are of special interest~\cite{choudhury2015robust}. Many new approaches have been studied to improve recognition performance, such as 3D body estimation~\cite{luo2016robust}, complete canonical correlation analysis~\cite{xing2016complete}, sparse coding and hypergraph learning~\cite{chen2016uncooperative}. However, mobile devices are becoming increasingly sophisticated and can provide high quality inertial measurements. Multiple activities can thus be analyzed using wearable sensors data, and exploited, e.g., for task identification~\cite{kwapisz2010cell}. A thorough review of the latest developments in this area can be found in Sprager's work~\cite{sprager2015inertial}.

Our interest in this paper is in human gait identification through smartphone inertial sensors. Ailisto et al.~\cite{mantyjarvi2005identifying} were the first to look at this problem and they did it through accelerometer data. In their paper, they used a triaxial accelerometer worn on a belt with fixed orientation: the \mbox{$x$-axis} pointed forward, the \mbox{$y$-axis} to the left and the \mbox{$z$-axis} was aligned with the direction of gravity. Only data points from the $x$ and $z$ axes were used for identification purposes. Gait cycle extraction was performed through a simple peak detection method, and a template was built for each subject. User identification employed a template matching technique, for which different methods were explored: temporal correlation, frequency domain analysis and data distribution statistics. 

In~\cite{derawi2010unobtrusive}, Derawi et al. proposed more robust preprocessing, cycle detection and template comparison algorithms. Data were acquired using a mobile phone worn on the hip, and only the vertical \mbox{$z$-axis} was considered for motion analysis. Dynamic Time Warping (DTW)~\cite{DTW-2005} was used as the distance measure, to ensure robustness against non-linear temporal shifts. This scheme was also tested in~\cite{nickel2011scenario}, where majority voting and cyclic rotation were compared as inference rules. In a further paper~\cite{nickel2011using}, Hidden Markov Models (HMM) were explored. Accelerometer data were split into windows of fixed length, which were then utilized to train the HMMs. Good identification results were obtained, but at the cost of long authentication phases ($30$ seconds).

Classification algorithms based on machine learning were also investigated. Either gait cycles extraction~\cite{juefei2012gait} or fixed windows lengths~\cite{nickel2012authentication} are possible signal segmentation methods. After that, a {\it feature extraction} technique is applied to each segment and statistical measures such as mean, standard deviation, root mean square, \mbox{zero-crossing} rate or histogram bin counts, are commonly used. However, more advanced features are required for better results, like cepstral coefficients, which are widely used for speech recognition~\cite{nickel2012authentication}, or features extracted through frequency analysis, i.e., using Fourier~\cite{thang2012gait} or wavelet transforms~\cite{juefei2012gait}. Supervised algorithms are typically used for classification, including \mbox{$k$-Nearest Neighbours} ($k$-NN)~\cite{choi2014biometric,nickel2012authentication,chan2013smart,Sprager-15}, Support Vector Machines (SVM)~\cite{juefei2012gait,chan2013smart,watanabe2014influence}, Multi Layer Perceptrons (MLP)~\cite{chan2013smart,watanabe2014influence} and Classification Trees (CT)~\cite{chan2013smart,watanabe2014influence}.

Accelerometer based gait analysis has also interest in the medical field. Using time-frequency analysis, Huang et al. showed that signals acquired by a waist-worn device on a patient with cervical disc herniation differed before and after the surgery~\cite{huang2012gait}. In~\cite{chan2013smart}, classification algorithms were used to discriminate a group of subjects with non-specific chronic low back pain from healthy subjects. Complex parameters, e.g., dynamic symmetry and cyclic stability of gait, were extracted by Jiang et al.~\cite{jiang2013possibility}. However, their evaluation requires to place sensors on the legs, and fine gait details are difficult to extract from signals acquired by a single waist-worn sensor.

We stress that in most of the related work the acquisition system was placed according to a {\it controlled and well known orientation} on the subject body. In real scenarios, this is however unlikely to be the case. It is thus important to implement an algorithm whose performance is invariant to the smartphone orientation, which is somewhat unconstrained (and unknown). 
This makes the phone reference system with good probability misaligned with respect to the direction of motion and the definition of subject specific and time invariant templates impossible. To deal with this, two different approaches can be used. The first consists in the extraction of features that are rotation invariant (e.g., correlation matrices of Fourier transforms~\cite{kobayashi2011rotation} or gait dynamic images~\cite{zhong2014sensor}). The second relies on the transformation of inertial signals~\cite{watanabe2014influence}, projecting them into a new {\it orientation invariant} three-dimensional reference system, which is extracted directly from the data. Here, we adopt the latter approach. Another distintive feature of our work is that we use an original processing pipeline exploiting automatic feature extraction through convolutional neural networks and a scoring algorithm combining one-class support vector machines and multi-step decision analysis. 

\begin{figure*}[tbp]
	\centering
	\includegraphics[width=\textwidth]{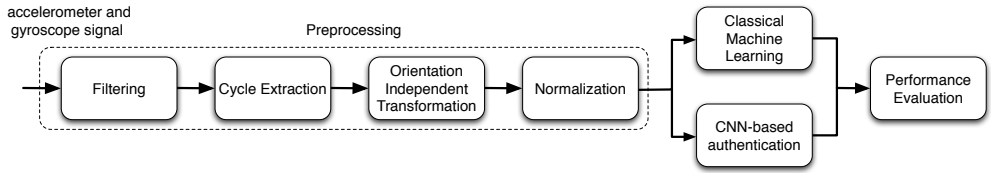}
	\caption{Signal processing workflow.}
	\label{fig:processing_workflow}
\end{figure*}




\section{Signal Processing Framework}
\label{sec:signal_acquisition_and_processing}

The aim of \fraName \, is to correctly recognize a subject from his/her way of walking, through the acquisition of inertial signals from a standard  smartphone.
The proposed processing workflow is shown in \fig{fig:processing_workflow}. Walking data is first acquired, then we perform some preprocessing entailing: 
\ben
\item pre-filtering to remove motion artifacts (\sect{sec:data_acquisition}),
\item the extraction of walking cycles (\sect{sec:cycle_extraction}), 
\item a transformation to move the raw walking data into a new {\it orientation independent} reference system (\sect{sec:reference_system_transformation}),
\item a normalization to represent each walking cycle (accelerometer and gyroscope data) through fixed length, zero mean and unit variance vectors (\sect{sec:normalization}). 
\een
After this, the walking cycles are ready to be processed to identify the user. The standard approach, called ``Classical Machine Learning'' entails the computation of a number of pre-established statistical features, the most informative of which are selected and used to train a classifier. Various machine learning techniques are usually exploited to this purpose, and are trained through a supervised approach. Hence, the classification performance is assessed and the whole process is usually iterated for a further feature selection phase. In this way, the features that are used for the classification task are progressively refined until a final feature set is attained. Note that statistical features are often assessed by the designer through educated guesses and a trial and error approach.

As opposed to this, with \fraName\ we advocate the use of convolutional neural networks (see Sections \ref{sec:multi-class-classification} and \ref{sec:one_class_classification}). These have been successfully used by the video processing community~\cite{CNN_video_analysis_2014} but to the best of our knowledge have never been exploited for the analysis of inertial data from wearable devices. One of the main advantages of this approach is that statistical features are automatically assessed by the CNN as a result of a supervised training phase. In \sect{sec:multi-class-classification}, the CNN is trained to act as a universal feature extractor, whereas in \sect{sec:one_class_classification} a one-class SVM  is trained as the final classifier. Once the CNN is trained, our system operates assuming that the smartphone only has access to the walking patterns of the target user (i.e., the legitimate user) and the SVM is solely trained using his/her walking data. Our system is based on the premise that, at runtime, the CNN should be capable of producing discriminant features for unseen users and the one-class SVM, once trained on the target, should reliably detect impostors, although their walks were not used for training.

The processing blocks are described in higher details in the following subsections.\\

\noindent \textbf{Notation:} With $\pmb{x} \in \mathbb{R}^n$ we mean a column vector $\pmb{x}=(x_1,x_2,\dots,x_n)^T$ with elements $x_i \in \mathbb{R}$, where $(\cdot)^T$ is the transpose operator. $|\pmb{x}|=n$ returns the number of elements in vector $\pmb{x}$. $\overline{x}=(\sum_{i=1}^n x_i)/n$, whereas $\|\pmb{x}\|=(\sum_{i=1}^n x_i^2)^{1/2}$ is the L2-norm operator. If $\pmb{x}, \pmb{y} \in \mathbb{R}^n$, we define their inner product as $\pmb{x} \cdot \pmb{y} = \pmb{x}^T \pmb{y}$ and their entrywise product as $\pmb{x} \circ \pmb{y} = (x_1 y_1, x_2 y_2, \dots, x_n y_n)^T$. Vector $\pmb{1}_n=(1,1,\dots,1)^T$ with $| \pmb{1}_n |=n$. Matrices are denoted by uppercase and bold letters. For example, if $\pmb{x}, \pmb{y}, \pmb{z} \in \mathbb{R}^n$, we define a $3 \times n$ matrix as $\pmb{M} = [\pmb{x}, \pmb{y}, \pmb{z}]^T$, whose rows contain the three vectors. In addition, element $(i,j)$ of matrix $\pmb{X}$ is denoted by $X_{i,j} \in \mathbb{R}$. With $\vec{\pmb{r}}$ we mean a 3D vector $\vec{\pmb{r}}=(r_1,r_2,r_3)^T$ and $\hat{\pmb{r}}$ is the corresponding 3D versor $\hat{\pmb{r}}=\vec{\pmb{r}}/\|\vec{\pmb{r}}\|$. For any two 3D vectors $\vec{\pmb{r}}$ and $\vec{\pmb{s}}$ we indicate their cross-product as $\vec{\pmb{r}} \times \vec{\pmb{s}} = (r_2 s_3 - r_3 s_2, r_3 s_1 - r_1 s_3, r_1 s_2 - r_2 s_1)^T$. The gravity vector is referred to as $\vec{\pmb{\rho}}$.  With $u(i)$ we mean a time series, where $i=1,2,\dots$ is the discrete time index. For acceleration data, the boldface letter $\pmb{a}$ is reserved for vectors, $a(i)$ for time series and $\pmb{A}$ for matrices. The same notation is adopted for the gyroscope data, using $\pmb{g}$, $g(i)$ and $\pmb{G}$, respectively for vectors, time series and matrices. 


\subsection{Data Acquisition and Filtering}
\label{sec:data_acquisition}

A proper dataset is key to the successful design and testing of identity recognition algorithms. Some datasets are publicly available. The largest one was acquired by the Institute of Scientific and Industrial Research (ISIR) at Osaka University (OU)~\cite{ngo2014largest}. It contains motion data collected from $744$ subjects using four motion sensors: three inertial sensors were placed on a belt, with triaxial accelerometer and gyroscope, and a smartphone was worn in the center back waist, and only measured triaxial accelerometer data. Despite the high number of participants, the main problem with this dataset is that motion data was acquired in a controlled environment, and for each subject only two short data sequences are available, which are not enough for deep network training. Furthermore, smartphone's gyroscope data is not provided. Other datasets are available, but featuring a much smaller number of participants. Casale et al. collected accelerometer data from a smartphone positioned in the chest pocket from $22$ users walking over a predefined path~\cite{casale2012personalization}. In~\cite{tilmanne2008adatabase}, a motion capture suit was used to acquire data from $40$ subjects walking in a small area at different speeds and with direction changes. However, due to the acquisition environment and conditions, these data are more suitable for human gait modeling rather than for user identification. Frank et al. collected data from a mobile phone in the pocket of $20$ individuals at McGill University, performing two separate $15$ minute walks on two different days~\cite{frank2010data}. Also in this case gyroscope data is not provided. All these databases do not meet our requirements. In fact, for proper training we need long data collection phases, preferably from different days and with devices freely worn in the user's front pockets. Hence, we decided to acquire our own motion traces, which are publicly available at \url{http://signet.dei.unipd.it/human-sensing/}.


Specifically, we acquired motion data from $50$ subjects, during a period of six months using Android smartphones worn in the right front pocket of the users' trousers. The following devices were used: Asus Zenfone 2, Samsung S3 Neo, Samsung S4, LG G2, LG G4 and a Google Nexus 5. Several acquisition sessions of about five minutes were performed for each subject, in variable conditions, e.g., with different shoes and clothes. We asked each subject to walk as she/he felt comfortable with, to mimic real world scenarios. For the data acquisition, we developed an Android inertial data logger application, which saves accelerometer, gyroscope and magnetometer signals into non-volatile memory and then automatically transfers them to an Internet server for further processing. The magnetometer signal is not used for identification purposes. In general, \fraName\ can be used carrying the device in other positions but we remark that each requires a dedicated training. 

\begin{figure}[t]
	\centering
	\includegraphics[width=\figw]{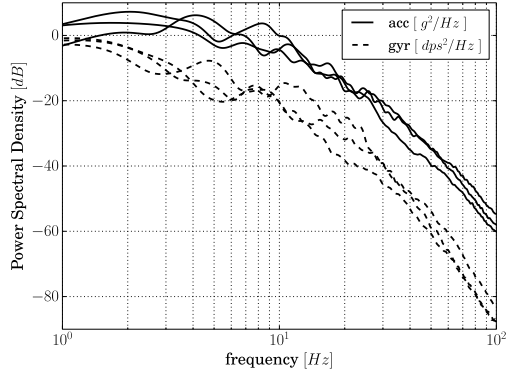}
	\caption{Power spectral density of accelerometer (continuous lines, one for each axis) and gyroscope (dashed lines) data.}
	\label{fig:psd}
\end{figure}

In \fig{fig:psd}, we plot the power of accelerometer and gyroscope signals at different frequencies through the Welch's method~\cite{Welch-67}, considering a full walking trace and setting the Hanning window length to $1$~s, with half window overlap. Most of the signal power is located at low frequencies, mostly below $40$~Hz (where the power is at least $30$~dB smaller than the maximum). The raw inertial signals were acquired using an average sample frequency ranging between $100$ and $200$~Hz (depending on the smartphone model), which is more than appropriate to capture most of the walking signal's energy.

At the first block of \fraName\ processing chain, due to the non-uniform sampling performed by the smartphone operating system, we apply a cubic Spline interpolation to represent the input data through evenly spaced points ($200$~points/second). Hence, a low pass Finite Impulse Response (FIR) filter with a cutoff frequency of $\cutoffone = 40$~Hz is used for denoising and to reduce the motion artifacts that may appear at higher frequencies. In fact, given the power profile of \fig{fig:psd}, the selected cutoff frequency only removes noise and preserves the relevant (discriminative) information about the user's motion.  

In the following, with $a_x(i)$ and $g_x(i)$ we respectively mean the filtered and interpolated acceleration and gyroscope time series along axis $x$, where $i=1,2,\dots$ is the sample number. The same notation holds for axes $y$ and $z$.

%
%


\subsection{Extraction of Walking Cycles} 
\label{sec:cycle_extraction}

\begin{figure}[!t]
	\centering
	\includegraphics[width=\figw]{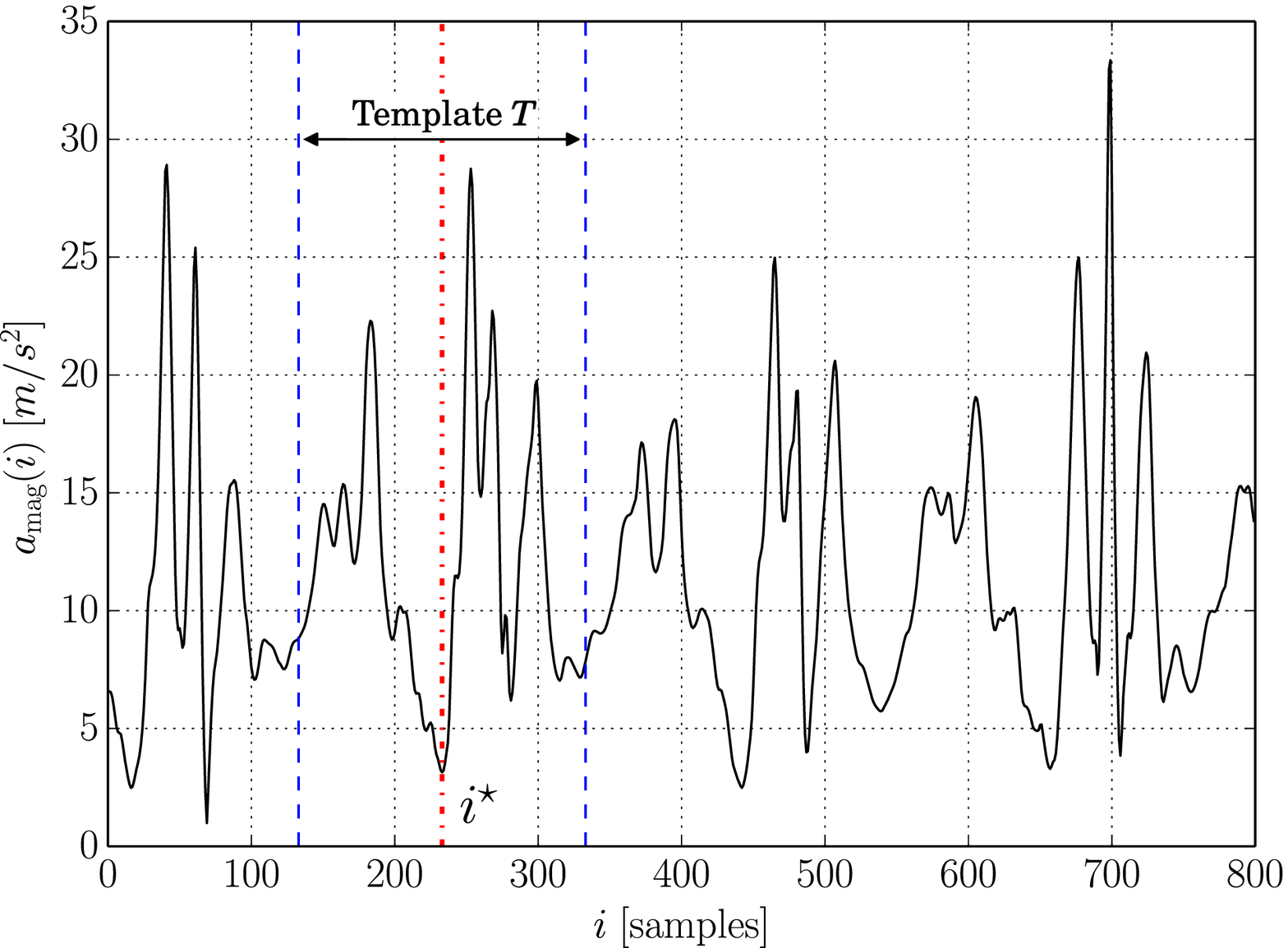}
	\caption{Template extraction using the accelerometer magnitude $a_{\rm mag}(i)$. The first template is the signal between the blue dashed vertical lines. The red dashed line in the center corresponds to $i^*$.}
	\label{img:first_template}
\vspace*{\floatsep}
	\centering
	\includegraphics[width=\figw]{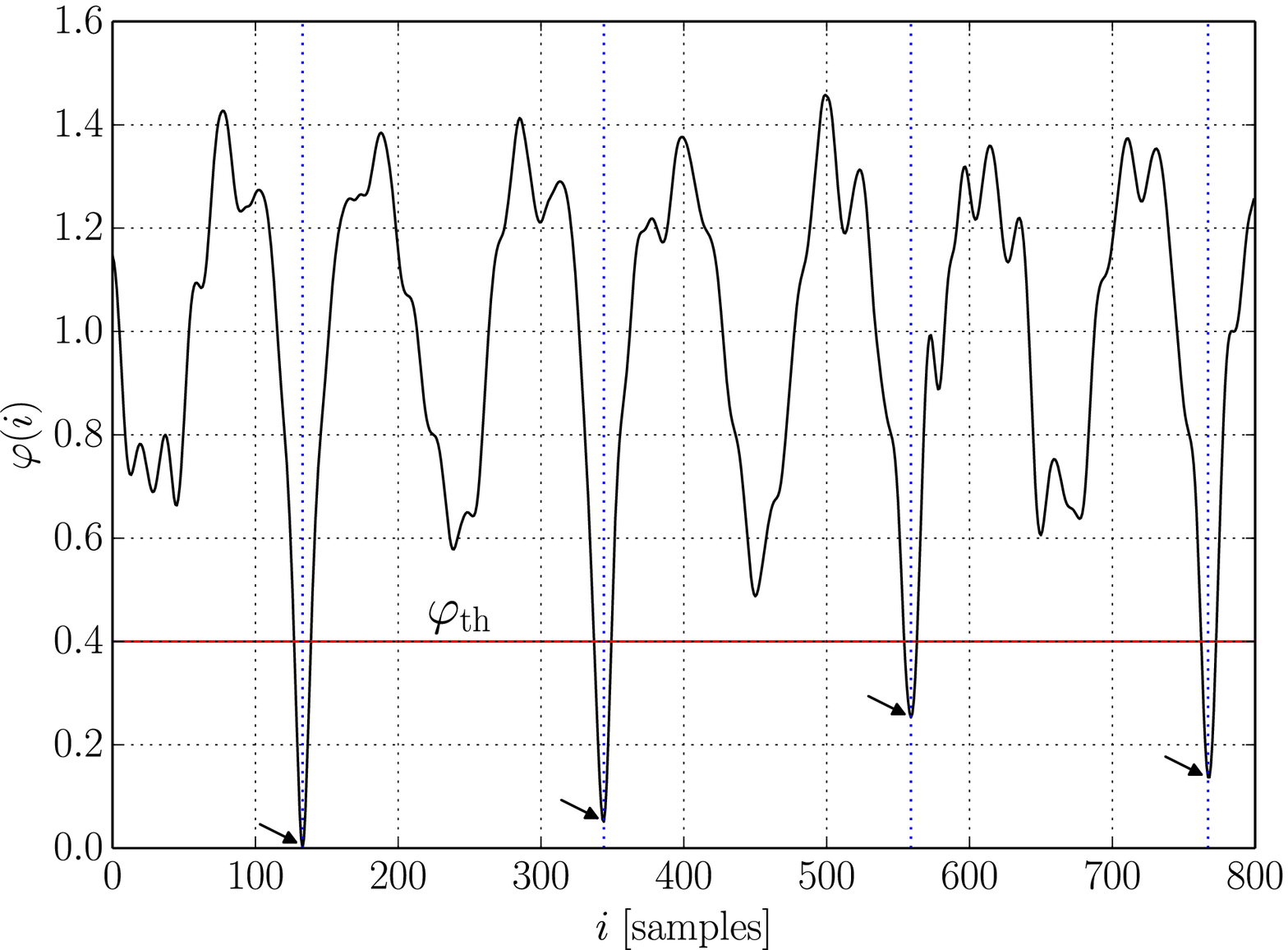}
	\caption{Correlation distance $\mycorr(i)$ between $a_{\rm mag}(i)$ and the template $\pmb{T}$ of \fig{img:first_template}. Local minima identify the beginning of  walking cycles.}
	\label{img:correlation_distance}
\end{figure}

For the extraction of walking cycles we use a template-based and iterative method that solely considers the accelerometer's magnitude signal. This signal is in fact inherently invariant to the rotation of the smartphone and, as such, allows for the precise assessment of walking cycles regardless of how the user carries the device in her/his front pocket. For each sample $i=1,2,\dots$ the acceleration magnitude is computed as:
\be
a_{\rm mag}(i) = (a_x(i)^2 + a_y(i)^2 + a_z(i))^{1/2} \, .
\ee
To identify the template, a reference point in $a_{\rm mag}(i)$ has to be located. To do so, inspired by~\cite{ren2013smartphone} we first pass $a_{\rm mag}(i)$ through a low-pass filter with cutoff frequency $\cutofftwo = 3$~Hz. Thus, we detect the first minimum of this filtered signal, which corresponds to the heel strike~\cite{teixeira2009distributed}, and the corresponding index is called $\tilde{i}$. This minimum is then refined by looking at the original signal $a_{\rm mag}(i)$ in a neighbourhood of $\tilde{i}$ and picking the minimum value of $a_{\rm mag}(i)$ in that neighborhood. This identifies a new index $i^\star$ for which $a_{\rm mag}(i^\star)$ is a local minimum. As an example, in \fig{img:first_template}, we show this minimum through a red vertical (dashed-dotted) line. As a second step, we pick a window of one second centered in $i^\star$, 
which in \fig{img:first_template} is represented through two vertical blue (dashed) lines. Now, the samples of $a_{\rm mag}(i)$ falling between the two blue lines define the first {\it gait template}, which we call $\pmb{T}$, with $|\pmb{T}| = \nsamples$~samples, where $\nsamples$ corresponds to the number of samples measured in one second.  

The extracted template is then iteratively refined and, at the same time, used to identify subsequent walking cycles. To this end, we first define the following correlation distance, for any two real vectors $\pmb{u}$ and $\pmb{v}$ of the same size $n$ we have:
\begin{equation}
\label{eq:correlation_distance}
\mathrm{corr\_dist}(\pmb{u},\pmb{v}) = 1 - \dfrac{(\pmb{u}-\overline{u}\pmb{1}_n) \cdot (\pmb{v}-\overline{v}\pmb{1}_n)}{\Vert(\pmb{u}-\overline{u}\pmb{1}_n)\Vert \  \Vert(\pmb{v}-\overline{v}\pmb{1}_n)\Vert} \, .
\end{equation}

The template $\pmb{T}$ is then processed with the acceleration magnitude through the following \eq{eq:gamma}, leading to a further metric $\mycorr(i)$, where $i=1,2,\dots$ is the sample index:  
\begin{eqnarray}
\label{eq:gamma}
\mathrm{vrect}(a_{\rm mag}(i)) & = (a_{\rm mag}(i), \dots, a_{\rm mag}(i+\nsamples-1))^T  \\
\mycorr(i) & = \mathrm{corr\_dist}(\pmb{T}, \mathrm{vrect}(a_{\rm mag}(i)) \, , \, i=1,\dots \, . \nonumber 
\end{eqnarray}
As can be seen from \fig{img:correlation_distance}, the function $\mycorr(i)$ exhibits some local minima, which are promptly located by comparing $\mycorr(i)$ with a suitable threshold $\mycorr_{\rm th}$ and performing a fast search inside the regions where $\mycorr(i)<\mycorr_{\rm th}$. The indices corresponding to these minima determine the optimal alignments between the template $\pmb{T}$ and $a_{\rm mag}(i)$. In particular, the second of these identifies the beginning of the second gait cycle. From these facts we have that: 
\ben
\item the samples between the second and the third minima correspond to the second gait cycle. It is thus possible to locate accelerometer and gyroscope vectors for this walking cycle, which are respectively defined as: $\pmb{a}_x$, $\pmb{a}_y$, $\pmb{a}_z$ and $\pmb{g}_x$, $\pmb{g}_y$, $\pmb{g}_z$, still expressed in the $(x,y,z)$ coordinate system of the smartphone. We remark that the number of samples does not necessarily match the template length and usually differs from cycle to cycle, as it depends on the length and duration of walking steps. 
\item A second template $\pmb{T}^\prime$ is obtained by reading $\nsamples$ samples starting from the second minimum. 
\een
At this point, a new template is obtained through a weighted average of the old template $\pmb{T}$ and the new one $\pmb{T}^\prime$:
\be
\label{eq:template_filtering}
\pmb{T} = \alpha \pmb{T} + (1-\alpha) \pmb{T}^\prime \, ,
\ee
where for the results in this paper we used $\alpha=0.9$. The new template $\pmb{T}$ is then considered for the extraction of the next walking cycle and the procedure is iterated. Note that this technique makes it possible to obtain an increasingly robust template at each new cycle.

A template matching approach exploiting a similar rationale was used in~\cite{ren2013smartphone}, where the authors employed the Pearson product-moment correlation coefficient between template and $a_{\rm mag}(i)$. The main differences between \cite{ren2013smartphone} and our approach are: we obtain the template $\pmb{T}$ in a neighborhood of $i^\star$, using a fixed number of samples $\nsamples$, whereas they take the samples between two adjacent minima of $\mycorr(i)$ (which may then differ in size for different cycles). In \eq{eq:template_filtering}, a discrete-time filter is utilized to refine the template $\pmb{T}$ at each walking cycle, making it more robust against speed changes. In previous work~\cite{ren2013smartphone}, the template is instead kept unchanged up to a point when minima cannot be longer detected, and a new template is to be obtained.

Finally, a normalization phase is required to represent all the cycles through the same number of points $N$, as this is required by the following feature extraction and classification algorithms. Before doing this, however, a transformation of accelerometer and gyroscope signals is performed to express these inertial signals in a {\it rotation invariant} reference system, as described next.

\begin{figure*}[t]
	\centering
	\includegraphics[width=\textwidth]{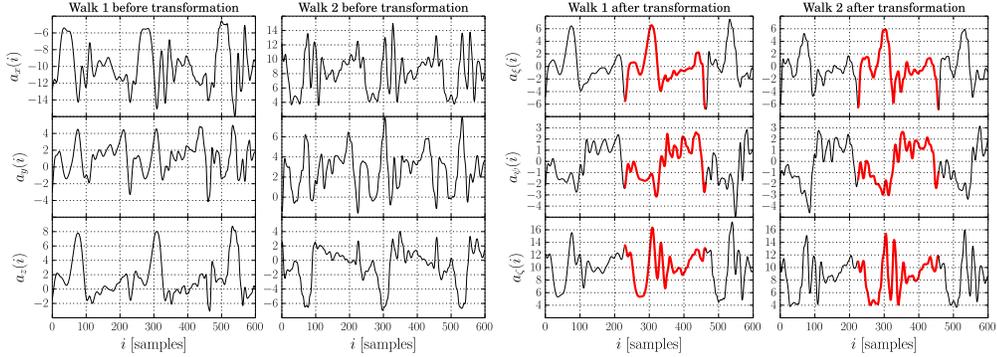}
	\caption{Raw accelerometer data from two different walks, acquired from a smartphone worn in the right front pocket with different orientations. Accelerometer data in the smartphone reference system $(x,y,z)$ (left), and after the transformation $(\xi,\psi,\zeta)$ (right). \fraName\ implements a PCA-based transformation that makes walking data rotation invariant, i.e., subject-specific gait patterns emerge in the new coordinate system (see the red colored patterns in the right plots).}
	\label{img:orientation_change}
\end{figure*}



\subsection{Orientation Independent Transformation}
\label{sec:reference_system_transformation}

To evaluate the new orientation invariant coordinate system, three orthogonal versors $\hat{\pmb{\xi}}$, $\hat{\pmb{\zeta}}$, $\hat{\pmb{\psi}}$ are to be found, whose orientation is independent of that of the smartphone and aligned with gravity and the direction of motion. Specifically, our aim is to express accelerometer and gyroscope signals in a coordinate system that remains fixed during the walk, with versor $\hat{\pmb{\zeta}}$ pointing up (and parallel to the user's torso), versor $\hat{\pmb{\xi}}$ pointing forward (aligned with the direction of motion) and $\hat{\pmb{\psi}}$ tracking the lateral movement and being orthogonal to the other two. This entails inferring the orientation of the mobile device carried in the front pocket from the acceleration signal acquired during the walk. To this end, we adopt a technique similar to those of~\cite{pca-heading-2009, heading-2015}.

Gravity is the main low frequency component in the accelerometer data, and will be our starting point for the transform. Moreover, although it is a constant vector, it  continuously changes when represented in the $(x,y,z)$ coordinate system of the smartphone, due to the user's mobility and the subsequent change of orientation of the device. So, even the gravity vector $\vec{\pmb{\rho}}$ is not constant when expressed through the smartphone coordinates. As the first axis of the new reference system, we consider the mean direction of gravity within the current walking cycle. Let $n_k$ be the number of samples in the current cycle $k$, with $k=1,2,\dots$. We recall that, with $\pmb{a}_x$, $\pmb{a}_y$ and $\pmb{a}_z$ we mean the acceleration samples in the current cycle $k$ along the three axes $x,y$ and $z$, with $|\pmb{a}_x|=|\pmb{a}_y|=|\pmb{a}_z|=n_k$, whereas with $\pmb{g}_x$, $\pmb{g}_y$ and $\pmb{g}_z$ we indicate the gyroscope samples in the same cycle $k$, with $|\pmb{g}_x|=|\pmb{g}_y|=|\pmb{g}_z|=n_k$. The gravity vector $\vec{\pmb{\rho}}_{k}$ within cycle $k$ is estimated as:
\begin{equation}
\vec{\pmb{\rho}}_{k} = ( \overline{a}_x , \overline{a}_y , \overline{a}_z )^T \, .
\label{eq:g_mean_vector}
\end{equation}
The first versor of the new system $\hat{\pmb{\zeta}}$ is obtained as:
\be
\hat{\pmb{\zeta}} = \dfrac{\vec{\pmb{\rho}}_{k}}{\Vert\vec{\pmb{\rho}}_{k}\Vert} \, .
\label{eq:gravity_versor1}
\ee
Now, we define the acceleration matrix $\pmb{A} = [\pmb{a}_x, \pmb{a}_y, \pmb{a}_z]^T$ of size $3 \times n_k$, whose rows corresponds to $\pmb{a}_x$, $\pmb{a}_y$ and $\pmb{a}_z$. Likewise, the gyroscope matrix is $\pmb{G} = [\pmb{g}_x, \pmb{g}_y, \pmb{g}_z]^T$, whose rows corresponds to $\pmb{g}_x$, $\pmb{g}_y$ and $\pmb{g}_z$. The projected acceleration and gyroscope vectors along axis $\hat{\pmb{\zeta}}$ are:
\begin{equation}
\pmb{a}_{\zeta} = \pmb{A} \cdot \hat{\pmb{\zeta}} \, , \, \pmb{g}_{\zeta} = \pmb{G} \cdot \hat{\pmb{\zeta}} \, ,
\label{eq:gravity_versor2}
\end{equation}
where the new vectors have the same size $n_k$. 
By removing this component from the original accelerometer signal, we project the latter on a plane that is orthogonal to $\hat{\pmb{\zeta}}$. This is the horizontal plane (parallel to the floor). We represent this {\it flattened} acceleration data through a new matrix $\pmb{A}^f = [\pmb{a}^f_{x}, \pmb{a}^f_{y}, \pmb{a}^f_{z}]^T$ of size $3 \times n_k$, where $\pmb{a}^f_{x}$, $\pmb{a}^f_{y}$ and $\pmb{a}^f_{z}$ are vectors of size $n_k$ that describe the acceleration on the new plane:
\be
\pmb{A}^{f} = \pmb{A} - \hat{\pmb{\zeta}} \pmb{a}_{\zeta}^T \, .
\label{eq:acc_flat}
\ee 
Analyzing this flattened acceleration, we see that during a walking cycle it is unevenly distributed on the horizontal plane. Also, the acceleration points on this plane are dispersed around a preferential direction, which has the highest excursion (variance). Here, we assume that the direction with the largest variance in our measurement space contain the dynamics of interest, i.e., it is parallel to the direction of motion, as it was also observed and verified in previous research~\cite{pca-heading-2009}. Given this, we pick this direction as the second axis (versor $\hat{\pmb{\xi}}$) of the new reference system. This is done by applying the Principal Component Analysis (PCA)~\cite{PCA-64} on the projected points, which finds the direction along which the variance of the measurements is maximized. The PCA procedure is as follows:

\ben
\item Find the empirical mean along each direction $x$, $y$ and $z$ (rows $1$, $2$ and $3$ of the flattened acceleration matrix $\pmb{A}^f$). Store the mean in a new vector $\pmb{u}$ of size $3 \times 1$., i.e.:
\be
u_i = \frac{1}{n_k} \sum_{j=1}^{n_k} A^f_{i,j}  \, , \, i=1,2,3 \, .
\ee
\item Subtract the empirical mean vector $\pmb{u}$ from each column of matrix $\pmb{A}^f$, obtaining the new matrix $\pmb{A}^f_{\rm norm}$:
\be
\pmb{A}^f_{\rm norm} = \pmb{A}^f - \pmb{u} (\pmb{1}_{n_k})^T \, .
\ee
\item Compute the sample $3 \times 3$ autocovariance matrix $\pmb{\Sigma}$:
\be
\pmb{\Sigma} = \frac{ \pmb{A}^{f}_{\rm norm} (\pmb{A}^{f}_{\rm norm})^T }{n_k-1} \, .
\label{eq:cov_mat}
\ee
\item The eigenvalues and the corresponding eigenvectors of $\pmb{\Sigma}$ are evaluated. The eigenvector $\vec{\pmb{v}}$ associated with the maximum eigenvalue identifies the direction of maximum variance in the dataset (i.e., the first principal component of the PCA transform).
\een


Hence, versor $\hat{\pmb{\xi}}$ is evaluated as:
\be
\hat{\pmb{\xi}} = \dfrac{\vec{\pmb{v}}}{\Vert\vec{\pmb{v}}\Vert} \, .
\label{eq:xi_versor1} 
\ee
Accelerometer and gyroscope data are then projected along $\hat{\pmb{\xi}}$ through the following equations: $\pmb{a}_{\xi} = \pmb{A} \cdot \hat{\pmb{\xi}}$ and $\pmb{g}_{\xi} = \pmb{G} \cdot \hat{\pmb{\xi}}$. Being $\hat{\pmb{\xi}}$ placed on a plane that is orthogonal to $\hat{\pmb{\zeta}}$, these two versors are also orthogonal. The third axis is then obtained through a cross product:
\be
\hat{\pmb{\psi}} = \hat{\pmb{\zeta}} \times \hat{\pmb{\xi}} \, , 
\ee
and the new accelerometer and gyroscope data along this axis are respectively obtained as: $\pmb{a}_{\psi} = \pmb{A} \cdot \hat{\pmb{\psi}}$ and $\pmb{g}_{\psi} = \pmb{G} \cdot \hat{\pmb{\psi}}$. The transformed vectors $(\pmb{a}_{\xi}, \pmb{a}_{\psi}, \pmb{a}_{\zeta})$ and $(\pmb{g}_{\xi}, \pmb{g}_{\psi}, \pmb{g}_{\zeta})$, along with the magnitude vectors $\pmb{a}_{\mathrm{mag}}$ and $\pmb{g}_{\mathrm{mag}}$ are the output of the Orientation Independent Transformation block of \fig{fig:processing_workflow}.

\begin{figure*}[t]
	\centering
	\includegraphics[width=\textwidth]{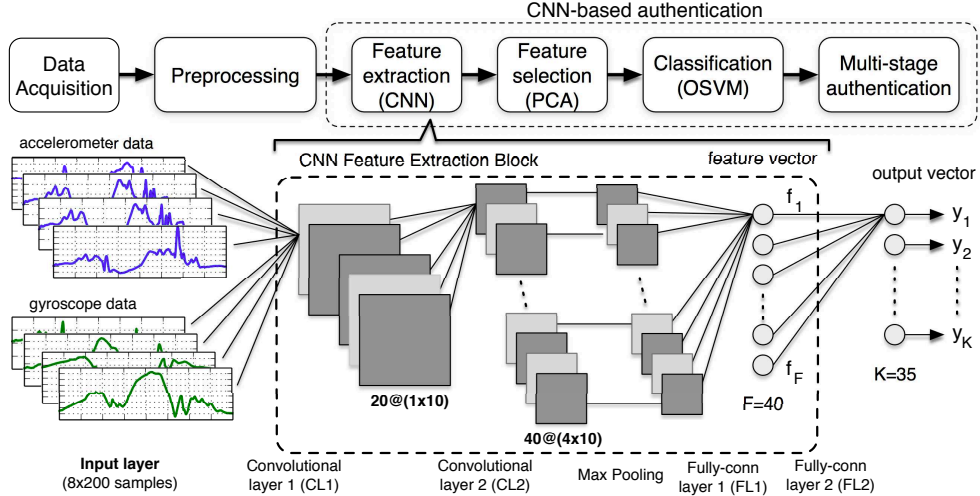}
	\caption{\fraName\ authentication framework. CL1 and CL2 are convolutional layers, FL1 and FL2 are fully connected layers. X@(Y$\times$Z) indicates the number of kernels, X, and the size of the kernel matrix, Y$\times$Z.}
	\label{fig:CNN_architecture}
\end{figure*}

An example of this transform is shown in \fig{img:orientation_change}, where accelerometer and gyroscope data from two different walks from the same subject are plotted. These signals were acquired carrying the phone in the right front pocket of the subject's trousers using two different orientations. As highlighted in the figure, our transform makes walking data rotation invariant. In fact, subject-specific gait patterns emerge in the new coordinate system (see the red colored patterns in the right plots).


\subsection{Normalization}
\label{sec:normalization}


Each gait cycle has a different duration, which depends on the walking speed and stride length. So, considering the accelerometer and gyroscope data collected during a full walking cycle, we remain with variable-size acceleration and gyroscope vectors, which are now expressed in the new orientation invariant coordinate system discussed in Section~\ref{sec:reference_system_transformation}. However, since feature extraction and classification algorithms require $N$-sized vectors for each cycle, where $N$ has to be fixed, a further adjustment is necessary. We cope with this cycle length variability through a further Spline interpolation to represent all walking cycles through vectors of $N=200$ samples each. This specific value of $N$ was selected to avoid aliasing. In fact, assuming a maximum cycle duration of $\tau=2$~seconds, which corresponds to a very slow walk, and a signal bandwidth of $B=40$~Hz, a number of samples $N > 2 B \tau = 160$~samples/cycle is required. Amplitude normalization was also implemented, to obtain vectors with zero mean and unit variance, as this leads to better training and classification performance. This results in a total of eight \mbox{$N$-sized} vectors for each walking cycle, which are inputted into the feature extraction and classification algorithms of the following sections.  



\section{Convolutional Neural Network}
\label{sec:multi-class-classification}

In this section, we present the chosen Convolutional Neural Network (CNN) architecture for \fraName\ (\sect{sec:conv_neural_net}), its optimization, training and quantitative comparison against gait authentication techniques from the literature (\sect{sec:network_optimization}).

\subsection{CNN Architecture}
\label{sec:conv_neural_net}
CNNs are feed-forward deep neural networks differing from fully connected multilayer networks for the presence of one or more convolutional layers. At each convolutional layer, a number of {\it kernels} is defined. Each of them has a number of weights, which are convolved with the input in a way that the same set of weights, i.e., the same kernel, is applied to all the input data, moving the convolution operation across the input span. Note that, as the same weights are reused  (shared weights), and each kernel operates on a small portion of the input signal, it follows that the network connectivity structure is sparse. This leads to advantages such as a considerably reduced computational complexity with respect to fully connected feed forward neural networks. For more details the reader is referred to~\cite{lecun1998convolutional}. CNNs have been proven to be excellent feature extractors for images~\cite{krizhevsky2012imagenet} and here we prove their effectiveness for motion data.
The CNN architecture that we designed to this purpose is shown in Fig.~\ref{fig:CNN_architecture}. It is composed of a cascade of two convolutional layers, followed by a pooling and a fully-connected layer. The convolutional layers perform a dimensionality reduction (or feature extraction) task, whereas the fully-connected one acts as a classifier. Accelerometer and gyroscope data from each walking cycle are processed according to the algorithms of \sect{sec:signal_acquisition_and_processing}. We refer to the input matrix for a generic walking cycle to as $\pmb{X} = (\pmb{a}_{\xi}, \pmb{a}_{\psi}, \pmb{a}_{\zeta}, \pmb{a}_{\mathrm{mag}}, \pmb{g}_{\xi}, \pmb{g}_{\psi}, \pmb{g}_{\zeta}, \pmb{g}_{\mathrm{mag}})^T$, where all the vectors are normalized to $N$ samples (see \sect{sec:normalization}). 
In detail, we have (CL~=~Convolutional Layer, FL~=~Fully-connected Layer):
\begin{itemize}[leftmargin=*]
	\item \textbf{CL1} The first convolutional layer implements one dimensional kernels ($1$x$10$ samples) performing a first filtering of the input and processing each input vector (rows of $\pmb{X}$) separately. This means that at this stage we do not  capture any correlation among different accelerometer and gyroscope axes. The activation functions are linear and the number of convolutional kernels is referred to as $\nkernels{1}$.
	\item \textbf{CL2} With the second convolutional layer we seek discriminant and class-invariant features. Here, the cross-correlation among input vectors is considered (kernels of size $4$x$10$ samples) and the output activation functions are non-linear hyperbolic tangents. Max pooling is applied to the output of CL2  to further reduce its dimensionality and increase the spatial invariance of features~\cite{scherer2010evaluation}. With $\nkernels{2}$ we mean the number of convolutional kernels used for CL2. 
	\item \textbf{FL1} This is a fully connected layer, i.e., each output neuron of CL2 is connected to all input neurons of this layer (weights are not shared). Hyperbolic tangent activation functions are used at the output neurons. FL1 output vector is termed $\pmb{f} = (f_1,\dots,f_F)^T$, and contains the $F$ features extracted by the CNN.
	\item \textbf{FL2} Each output neuron in this layer corresponds to a specific class (one class per user), for a total of $K$ neurons, where $K$ is the number of subjects considered for the training phase. The $K$ dimensional output vector $\pmb{y} = (y_1,\dots,y_K)^T$ is obtained by a \textit{softmax} activation function, which implies that $y_j \in (0,1)$, $j=1,\dots,K$ and $\sum_{j=1}^K y_j = 1$ (stochastic vector). Also, $y_j$ can be thought of as the probability that the current data matrix $\pmb{X}$ belongs to class (user) $j$.
\end{itemize}
%
The network is trained in a supervised manner for a total of $K$ subjects solving a multi-class classification problem, where each of the input matrices $\pmb{X}$ in the dataset is assigned to one of $K$ mutually exclusive classes. The {\it target} output vector $\pmb{t} = (t_1,\dots,t_K)^T$ has binary entries and is encoded using a \mbox{$1$-of-$K$} coding scheme, i.e., they are all zero except for that corresponding to the subject that generated the input data.



\subsection{CNN Optimization and Results}
\label{sec:network_optimization}

In this section, we propose some approaches for the optimization of the CNN, quantify its classification performance and compare it against classification techniques from the literature. As said above, the output of layer FL2 is the stochastic vector $\pmb{y}$, whose $j$-th entry $y_j$, $j=1,\dots,K$, can be seen as the probability that the input pattern belongs to user $j$, i.e., $y_j = y_j(\pmb{w}, \pmb{X}) = \textrm{Prob}(t_j = 1 | \pmb{w}, \pmb{X})$, where $\pmb{w}$ is the vector containing all the CNN weights, $\pmb{X}$ is the current input matrix (walking cycle) and $t_j=1$ if $\pmb{X}$ belongs to class $j$ and $t_j=0$ otherwise. If $\mathcal{X}$ is the set of all training examples, we define the batch set as $\mathcal{B} \subset \mathcal{X}$. Let $\pmb{X} \in \mathcal{B}$ and denote the corresponding output vector by $\pmb{y}(\pmb{w}, \pmb{X})$ and its $j$-th entry by $y_j(\pmb{w}, \pmb{X})$. The corresponding target vector is $\pmb{t}(\pmb{X}) = (t_{1}(\pmb{X}),\dots,t_K(\pmb{X}))^T$. 
The CNN is then trained through a stochastic gradient descend algorithm which minimizes a {\it categorical cross-entropy loss function} $L(\pmb{w})$, defined as \cite[Eq. (5.24) of Section~5.2]{Bishop-2007}:
\begin{equation}
L(\pmb{w}) = - \sum_{\pmb{X} \in \mathcal{B}} \sum_{j=1}^K t_{j}(\pmb{X}) \log(y_j(\pmb{w}, \pmb{X})) \, .
\label{eq:loss_function} 
\end{equation}
During training, \eq{eq:loss_function} is iteratively minimized, by rotating the walking cycles (training examples) in the batch set $\mathcal{B}$ so as to span the entire input set $\mathcal{X}$. Training continues until a stopping criterion is met (see below).


Walking patterns from $K$ subjects are used to train the CNN, and the same number of cycles $\ncycles$ is considered for each of them, for a total of $K \ncycles$ training cycles. $\ntest$ randomly chosen walking cycles from each subjects are used to obtain a test set $\mathcal{P}$. The remaining cycles are split into training $\mathcal{T}$ and validation $\mathcal{V}$ sets, with $|\mathcal{P}|= K \ntest$, $|\mathcal{T}|=K \ncycles$, $\mathcal{X} = \mathcal{P} \cup \mathcal{T} \cup \mathcal{V}$, where all the sets have null pairwise intersection and are built picking input patterns from $\mathcal{X}$ evenly at random. Set $\mathcal{V}$ is used to terminate the training phase, and termination occurs when the loss function $L(\pmb{w})$ evaluated on $\mathcal{V}$ does not decrease for twenty consecutive training epochs. After that, the network weights which led to the minimum validation loss are used to assess the CNN performance on set $\mathcal{P}$. This is done through an \textit{accuracy} measure, defined as the number of walking cycles correctly classified by the CNN divided by the total number of cycles in $\mathcal{P}$. 
In the following graphs, we show the mean accuracy obtained averaging the test set performance over ten different networks, all of them trained through the just explained approach by considering $K=35$ subjects from our dataset and $\ntest=100$ cycles per subject.

\begin{figure}[!t]
	\centering
	\includegraphics[width=\figw]{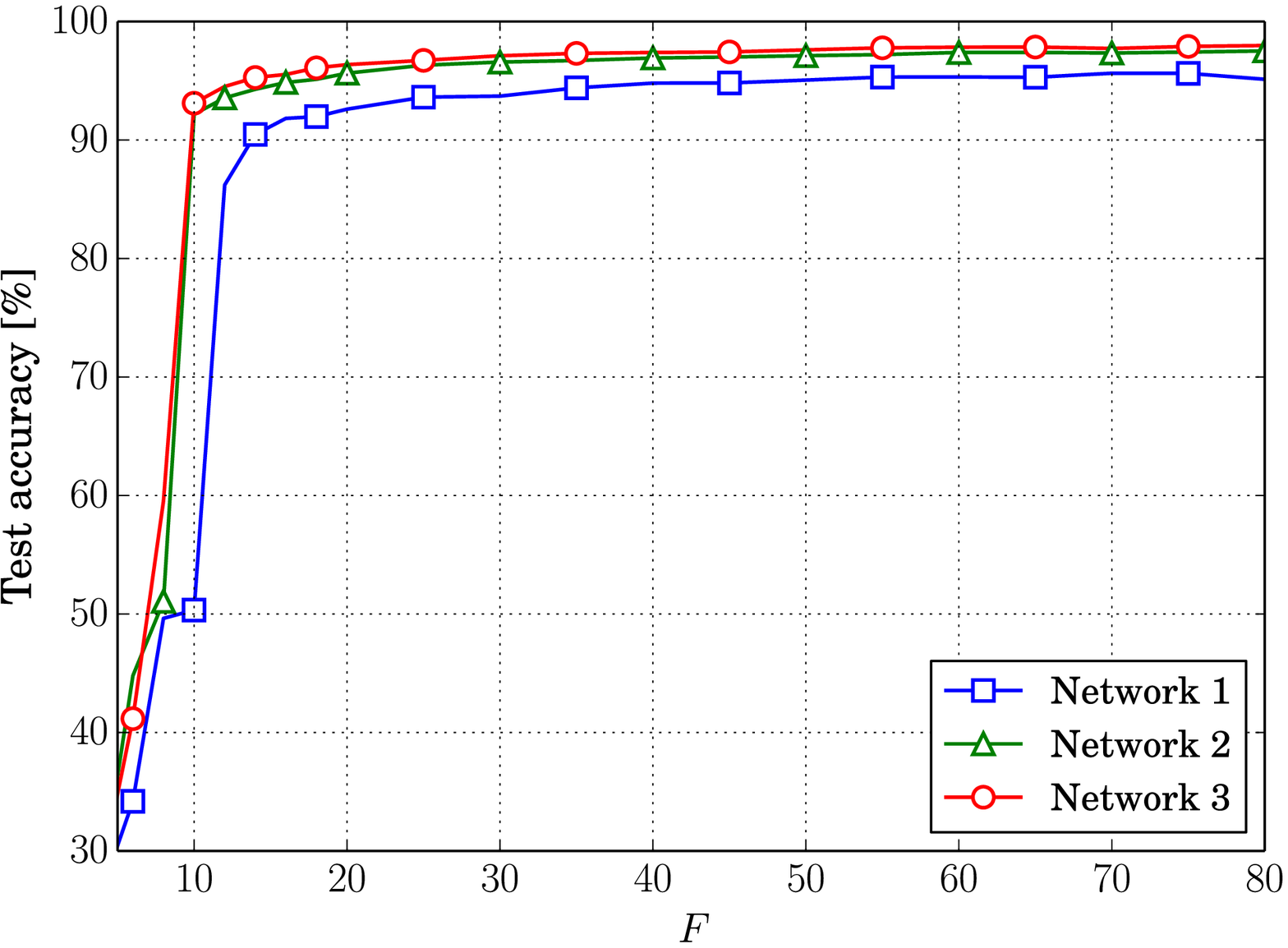}
	\caption{CNN test accuracy {\it vs} number of features $F$ in layer FL1. Three curves are shown for three different network configurations (number of kernels in layers CL1 and CL2).}
	\label{img:Nfeat_vs_acc}
\vspace*{\floatsep}
	\centering
	\includegraphics[width=\figw]{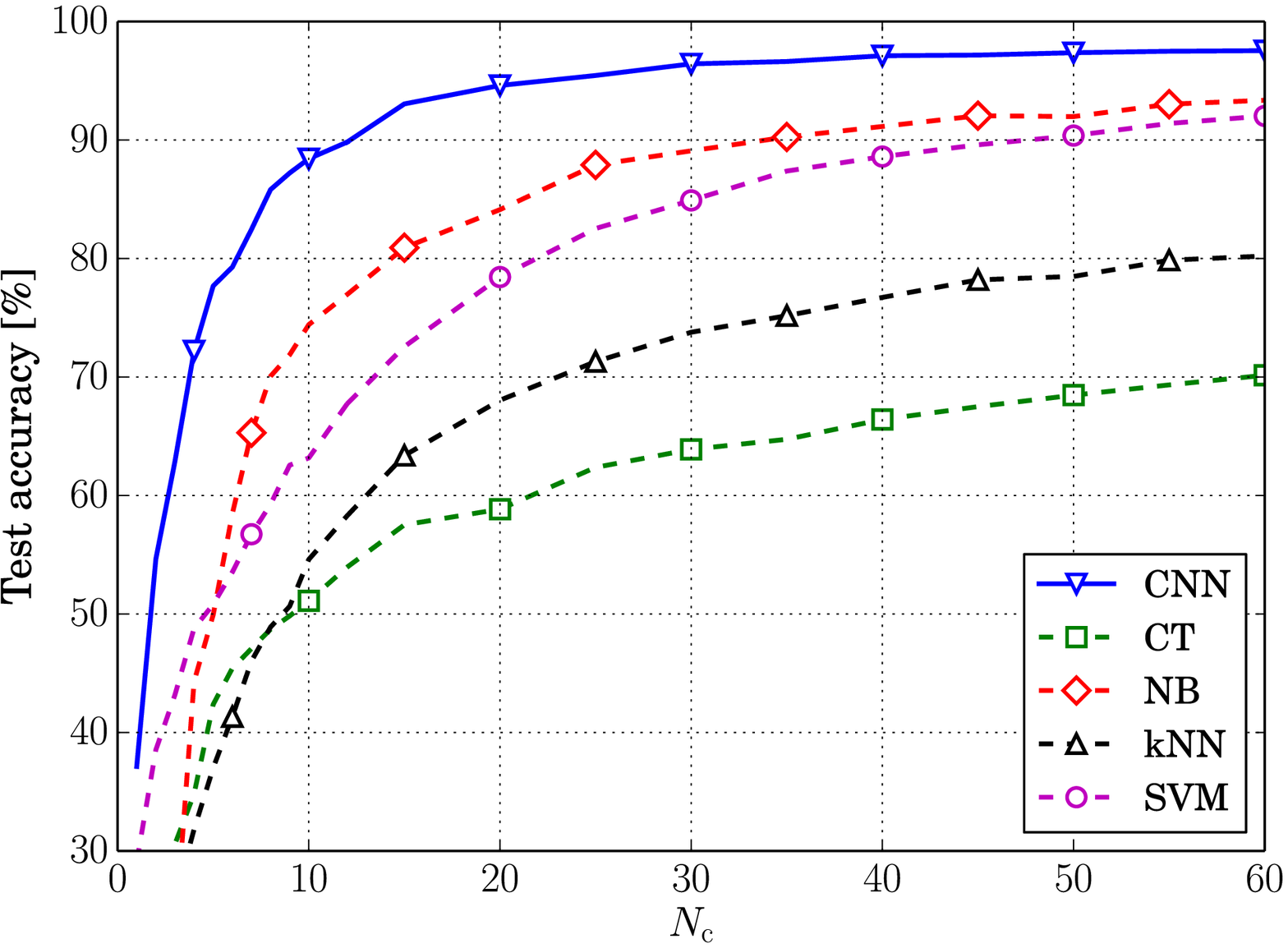}
	\caption{CNN test accuracy {\it vs} number of walking cycles $\ncycles$ used for training. Results for CT, NB, $k$-NN and SVM classifiers from the literature are also shown.}
	\label{img:ncyc_vs_acc}
\end{figure}

As a first set of results, we look at the impact of $F$ (neurons in layer $FL1$) and of the number of convolutional kernels in CL1 and CL2. Since the last layer FL2 acts as a classifier, $F$ can be seen as the number of features extracted by the CNN. In general, a too small $F$ can lead to poor classification results; too many features, instead, would make the state space too big to be effectively dealt with (curse of dimensionality)~\cite{hanka1997curse}. Besides $F$, we also investigate the right number of kernels to use within each convolutional layer. Three networks are considered by picking different $(\nkernels{1}, \nkernels{2})$ pairs. For network 1 we use $(\nkernels{1}=10, \nkernels{2}=20)$, network 2 has $(\nkernels{1}=20, \nkernels{2}=40)$ and network 3 has $(\nkernels{1}=30, \nkernels{2}=50)$. In \fig{img:Nfeat_vs_acc}, we show the accuracy performance of these networks as a function of $F$. From this plot, it can be seen that at least $F=20$ neurons have to be used at the output of FL1 and that the accuracy performance stabilizes around $F=40$, leading to negligible improvements as $N$ grows beyond this value. As for the number of kernels, we conclude that small networks (network 1) perform worse than bigger ones (networks 2 and 3), but increasing the number of kernels beyond that used for network 2 does not lead to appreciable improvements. Hence, for the results of this paper we used $F=40$ with $(\nkernels{1}=20, \nkernels{2}=40)$.

\textbf{Comparison against existing techniques:} in \fig{img:ncyc_vs_acc}, the accuracy is plotted against $\ncycles$ for our CNN-based approach and four selected authentication algorithms from the literature, featuring classifiers based on Classification Trees (CT)~\cite{quinlan1993c45}, Naive Bayes (NB)~\cite{friedman1997bayesian}, $k$-Nearest Neighbors ($k$-NN)~\cite{cover1967nearest} and Support Vector Machines (SVM)~\cite{cortes1995support}.\footnote{For SVM, we considered a linear kernel, as it outperformed polynomial and radial basis function ones (results are omitted in the interest of space). A one-versus-all strategy was used solve the considered multiclass problem for the binary classifers.} These techniques were used in a large number of papers including~\cite{choi2014biometric,nickel2012authentication,juefei2012gait,chan2013smart,watanabe2014influence}.
For their training, $112$ features were extracted from the signal samples in $\pmb{X}$, including their variance, mean trend, windowed mean difference, variance trend,  windowed variance difference, maxima and minima, spectral entropy, zero crossing rate and bin counts. These features, were then utilized to train the selected classifiers in a supervised manner. Note that, while the CNN automatically extracts its features (vector $\pmb{f}$), with previous schemes these are manually selected based on experience. 

From \fig{img:ncyc_vs_acc}, we see that the CNN-based algorithm delivers better accuracies across the entire range of $\ncycles$. Also, the accuracy increases with an increasing $\ncycles$ until it saturates and no noticeable improvements are observed. While a higher $\ncycles$ is always beneficial, a higher number of cycles also entails a longer acquisition time, which we would rather avoid. For this reason, for the following results we have used $\ncycles=40$ as it provided a good tradeoff between accuracy and complexity across all our experiments. 

\begin{figure}[!t]
	\centering
	\includegraphics[width=\figw]{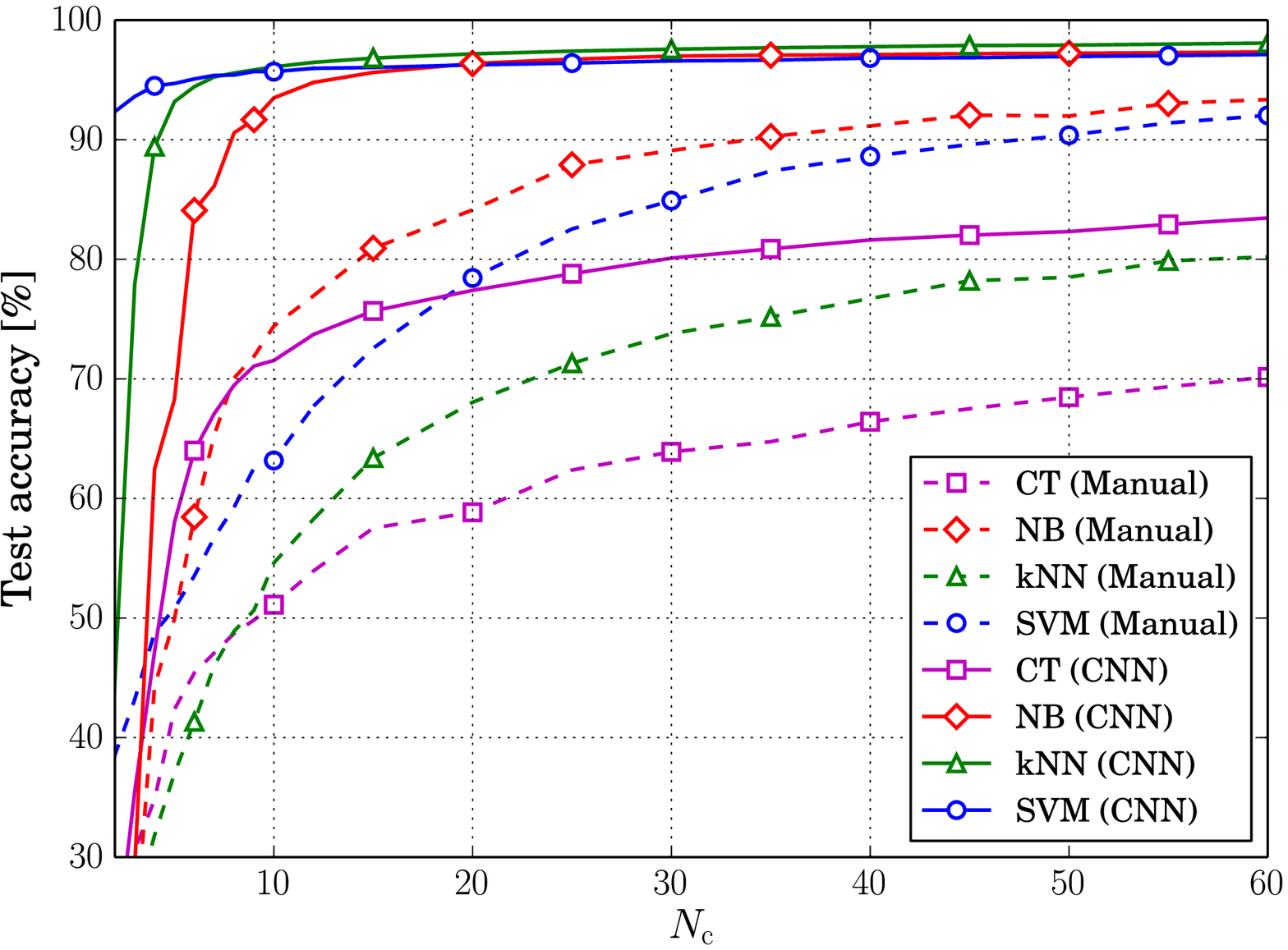}
	\caption{Test accuracy of CT, NB, $k$-NN and SVM classifiers. ``CNN'' indicates training with CNN-extracted features, whereas ``Manual'' means standard feature extraction.}
	\label{img:class_vs_CNN}
\vspace*{\floatsep}
	\centering
	\includegraphics[width=\figw]{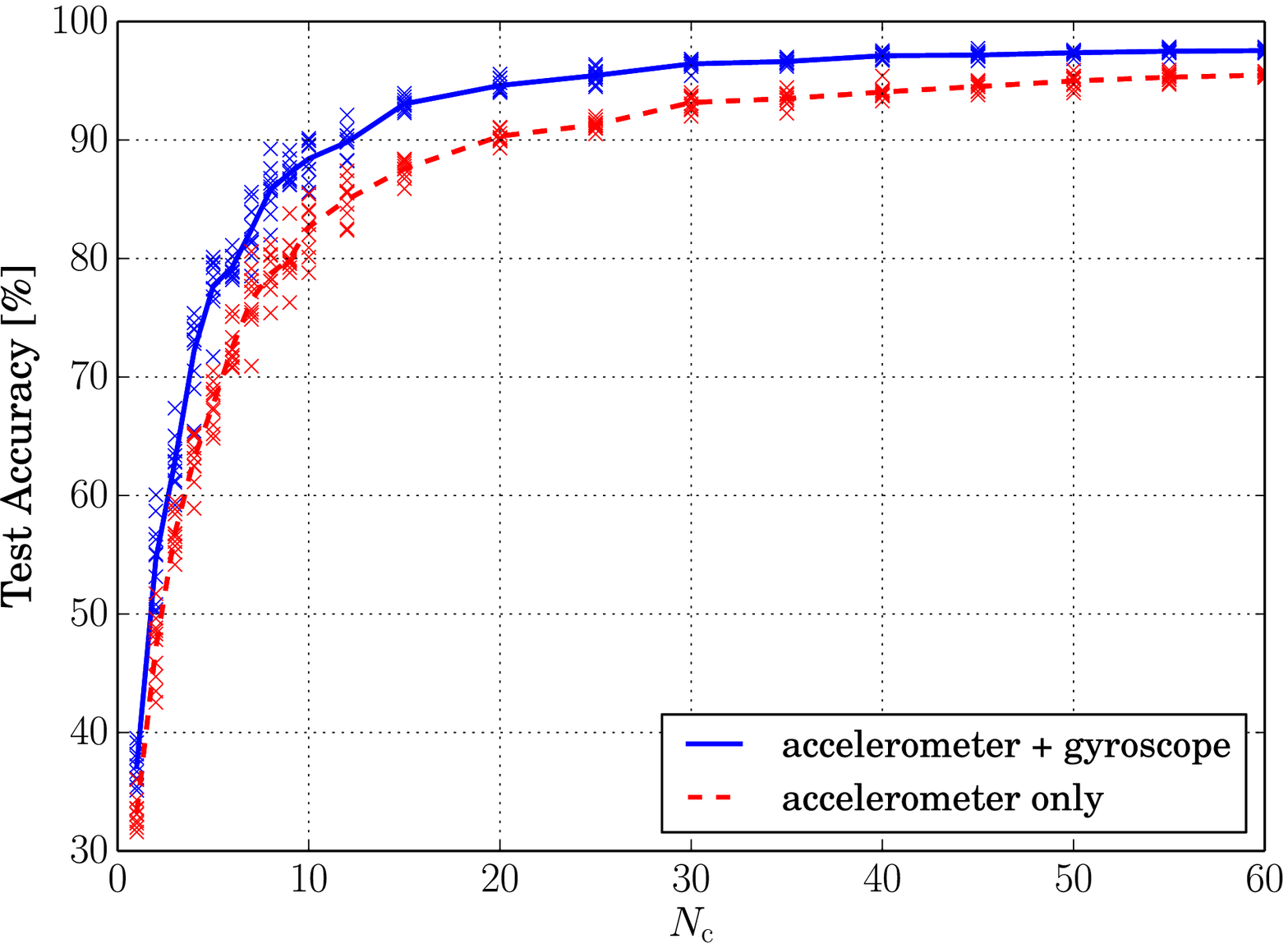}
	\caption{Impact of gyroscope data. Lines represent the mean accuracy (averaged over ten networks), whereas markers indicate the results of the ten network instances.}
	\label{img:with_without_gyro}
\end{figure}


To illustrate the superiority of CNN features with respect to manually extracted ones, in the following we conduct an instructive experiment. We consider CNN as a feature extraction block, by removing the output vector $\pmb{y}$ and using the inner feature vector $\pmb{f}$ to train the above classifiers from the literature (CT, NB, $k$-NN and SVM). The corresponding accuracy results are provided in \fig{img:class_vs_CNN}.  All the classifiers perform better when trained using CNN features, with typical improvements in the test accuracy of more than $10$\%. For instance, for a $k$-NN classifier trained with $\ncycles=30$ cycles per subject, the accuracy increases from $71$\% (manually extracted features) to $94\%$ (CNN features). The best performance is provided by the combined use of CNN features and SVM. 

A last consideration is in order. Most of the previous papers only used accelerometer data, but our results show that using both gyroscope and accelerometer provides further improvements, see \fig{img:with_without_gyro}.


\section{One-Class Support Vector Machine Training}
\label{sec:one_class_classification}

In this section, we further extend the \fraName\ CNN-based authentication chain through the design of an SVM classifier which is trained solely using the motion data of the target subject. This is referred to as One-Class Classification (OCC) and is important for practical applications where motion signals of the target user are available, but those belonging to other subjects are not. More importantly, with this approach the classification framework can be extended to users that were not considered in the CNN training. 

\subsection{Revised Classification Architecture}
\label{sec:revised_architecture}

Due to the generalization properties of convolutional deep networks, once trained, the CNN can be used as a universal feature extractor, providing meaningful features even for subjects that were not included in the training. To take advantage of this, we discard the output neurons of FL2 and utilize the CNN as a dimensionality reduction tool that, given an input matrix $\pmb{X}$, returns a user dependent feature vector $\pmb{f}$. The CNN is then trained only once considering the optimizations of \sect{sec:network_optimization}. All its weights and biases are then precomputed and will not be modified at classification time.
Considering the diagram of Fig.~\ref{fig:CNN_architecture}, at the output of the CNN we obtain the feature vector $\pmb{f}$. We then apply a feature selection block to reduce the number of features from $F$ to $S \leq F$ (dimensionality reduction). PCA is used to accomplish this task and the new feature vector is called $\pmb{s}$. Hence, we have $\pmb{s} = \PCA(\pmb{f})$, where $\PCA(\cdot) : \mathbb{R}^F \to \mathbb{R}^{S}$ is the PCA transform. 

A One-class Support Vector Machine (OSVM) is then used as the classification algorithm (\sect{sec:OSVM-classification}). It defines a {\it boundary} around the feature (training) vectors belonging to the target subject. At runtime, as a new walking cycle is processed, the OSVM takes the feature vector $\pmb{s}$ and outputs a {\it score}, which is a distance measure between the current feature vector and the SVM boundary~\cite[Chapter 7]{Bishop-2007}. As we discuss shortly, this score relates to the likelihood that the current walking cycle belongs to the target user. 

\subsection{One-Class SVM Design}
\label{sec:OSVM-classification}

Next, we design the OSVM block of Fig.~\ref{fig:CNN_architecture}. It differs from a standard binary SVM classifier as the SVM boundary is built solely using patterns from the positive class (target user). The strategy proposed by Sch\"olkopf is to map the data into the feature space of a kernel, and to separate them from the origin with maximum margin~\cite{scholkopf1999support}. The corresponding minimization problem is similar to that of the original SVM formulation~\cite{cortes1995support}. The trick is to use a hyperplane (in the space transformed by a suitable kernel function) to discriminate the target vectors. OSVM takes as input the reduced feature vector $\pmb{s} = (s_1, \dots, s_{S})^T$ and we use the following Radial Basis Function (RBF) kernel, that for any $\pmb{s}, \pmb{s}^\prime \in \mathbb{R}^S$ is defined as:
\begin{equation}
\SVMkernel(\pmb{s},\pmb{s}')  = \left( \SVMfeatmap(\pmb{s}) \cdot \SVMfeatmap(\pmb{s}') \right) = \text{exp} \left(-\gamma\ \| \pmb{s}-\pmb{s}' \|^2 \right) \, ,
\label{eq:rbf_kernel}
\end{equation}
where $\SVMfeatmap(\pmb{s})$ is a feature map and $\gamma$ is the RBF kernel parameter, which intuitively relates to the radius of influence that each training vector has for the space transformation. With $\ell$ we mean the number of training points (feature vectors), $\pmb{\omega}$ and $b$ are the  hyperplane parameters in the transformed domain (through \eq{eq:rbf_kernel}) and $\pmb{\varepsilon} = (\varepsilon_1, \dots, \varepsilon_\ell)^T$ is the vector of slack variables, which are introduced to deal with outliers. Given this, the following quadratic program is defined to separate the feature vectors in the training set, $\pmb{s}_1,\dots,\pmb{s}_\ell$, from the origin:
\begin{eqnarray}
\label{eq:svm_min_prob}
\min_{\pmb{\omega},\pmb{\varepsilon},b} \quad  &\dfrac{1}{2} \|\pmb{\omega}\|^2 + \dfrac{1}{\nu \ell}\sum_{j=1}^{\ell} \varepsilon_j - b  \\
\text{subject to} \quad &(\pmb{\omega} \cdot \Phi(s_j)) \geq b - \varepsilon_j \, , \, \varepsilon_j \geq 0 \, , \, j=1,\dots,\ell \nonumber 
\end{eqnarray}
$\nu \in (0,1)$ is one of the most important parameters and sets an upper bound on the fraction of outliers and a lower bound on the fraction of Support Vectors (SV)~\cite{scholkopf1999support}.
The decision function for a generic feature vector $\pmb{s}$ is defined as \mbox{$\decisionfunc(\pmb{s}) \in \{-1,+1\}$}, is obtained solving \eq{eq:svm_min_prob}, and only depends on the training vectors through the following relations:
\begin{equation}
\decisionfunc(\pmb{s}) = \text{sgn}\left( \scorefunc(\pmb{s}) \right) , \, \scorefunc(\pmb{s}) = \sum_{j=1}^\ell \alpha_j \SVMkernel(\pmb{s}_j,\pmb{s}) - b \, .
\label{eq:svm_dec_func}
\end{equation}
Now, $\alpha_j \geq 0, \forall \, j$, and only some of the training vectors have $\alpha_j >0$. These are the {\it support vectors} associated with the classification problem and are the only ones who count in the definition of the SVM boundary. $\scorefunc(\pmb{s})$ is the {\it score} associated with vector $\pmb{s}$. It weighs the distance from the SVM boundary, i.e., is greater than zero if $\pmb{s}$ resides inside the boundary, zero if it lies on it and negative otherwise.

Hence, the SVM is trained using a set of $\ell$ feature vectors from the target user, obtaining the SVM boundary (and the related decision function) through \eq{eq:svm_dec_func}. After training, we test the performance of the obtained SVM classifier considering feature vectors from the positive class $\positiveC$ (target user) and the negative one $\negativeC$ (any other user). Note that the vectors used for this test were not considered during the SVM training. 

As it is customary for binary classification approaches, the two most important metrics to assess the goodness of a classifier are the {\it precision} and the {\it recall}. The precision is the fraction of true positives, i.e., the fraction of patterns identified of the target class that in fact belong to the target user, while the recall corresponds to the fraction of target patterns that are correctly classified out of the entire positive class of samples~\cite{Fmeasure-2003}. Often, these two metrics are combined into their harmonic mean, which is called F-measure and is used as  the single quality parameter.    

\begin{figure}[t]
	\centering
	\includegraphics[width=\figw]{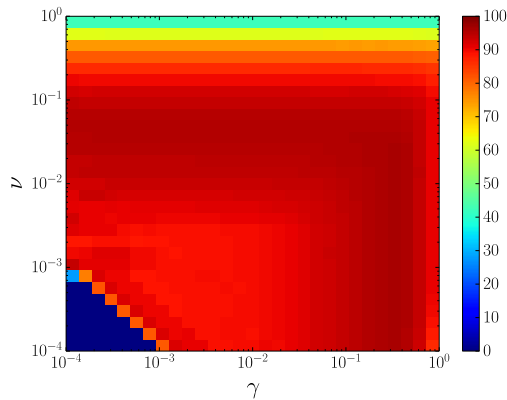}
	\caption{OSVM: F-measure as a function of $\gamma$ and $\nu$.}
	\label{fig:05_OSVM_nu_vs_gamma}
\end{figure}

In Fig.~\ref{fig:05_OSVM_nu_vs_gamma}, the F-measure is plotted as a function of the two SVM parameters $\gamma$ and $\nu$.  As seen from this plot, the area where the classifier's performance is maximum is quite ample. This is good as it means that even selecting $\gamma$ and $\nu$ once for all at design stage, the performance of the SVM classifier is not expected to change much if the signal statistics changes or a new target user is considered. In other words, this relatively weak dependence on the parameters entails an intrinsic robustness for the classifier. For the results that follow we have used $\gamma= 0.3$ and $\nu=0.02$. 

\begin{figure}[!t]
	\centering
	\includegraphics[width=\figw]{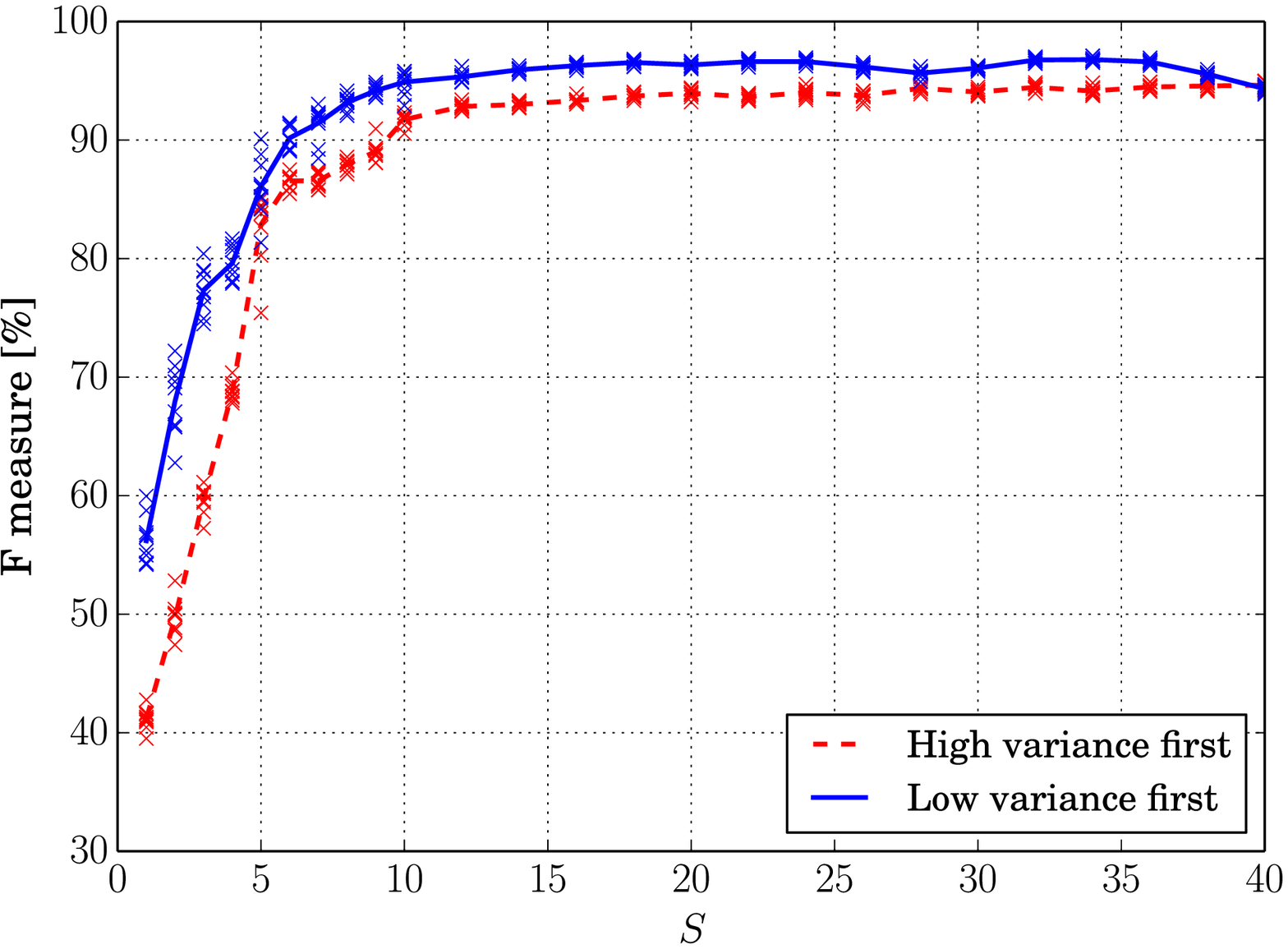}
	\caption{OSVM: F-measure as a function of the number of retained PCA features $S$. The number of CNN-extracted features is $F=40$.}
	\label{fig:07_OSVM_Npca_optimization}
\vspace*{\floatsep}
	\centering
	\includegraphics[width=\figw]{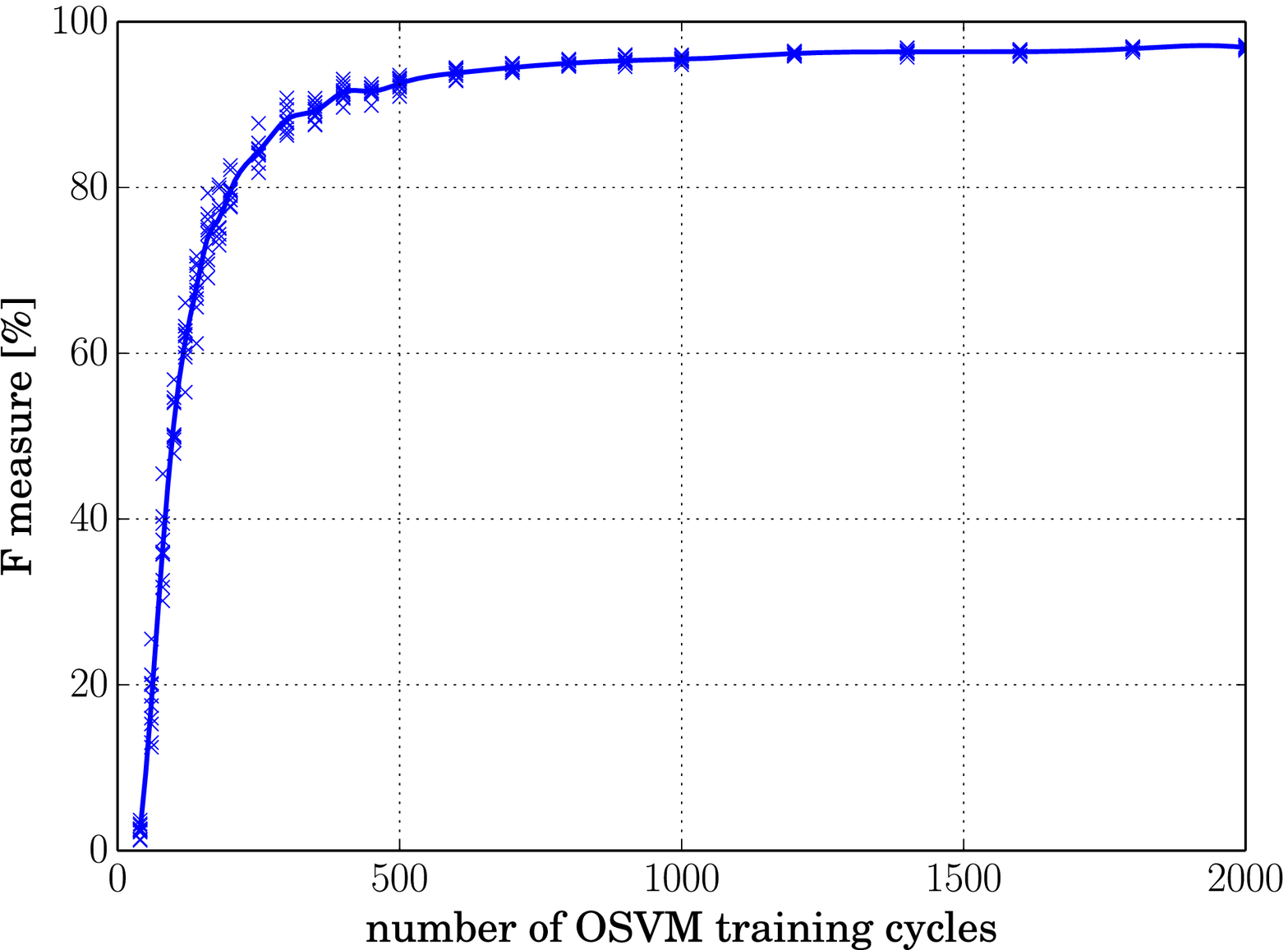}
	\caption{F-measure as a function of the number of walking cycles used to train the OVSM classifier. }
	\label{fig:06_OSVM_Ncyc_optimization}
\end{figure}

Two last considerations are in order. The first relates to the PCA transformation $\Upsilon(\cdot)$ and in particular to how many and which principal components have to be retained for the output feature vectors. In fact, as pointed out in~\cite{tax2003artificial}, two options are possible to go from the CNN-extracted feature vector $\pmb{f}$ to $\pmb{s}$. The first is to retain the $S \leq F$ entries of the transformed vector (expressed in the PCA basis) that correspond to the principal components with highest variance, whereas a second option is to retain those with the smallest. Fig.~\ref{fig:07_OSVM_Npca_optimization} shows the F-measure of the OSVM classifier as a function of $S$ for $F=40$ (number of CNN-extracted features). From this plot we see that picking $S<F$ in general provides better results and also that considering the principal components with lowest variance provides better results for this class of problems. This is in accordance with~\cite{tax2003artificial}.

The last consideration regards the amount of feature vectors belonging to the target user that should be used for the OSVM training. Note that this number is related to the walking time required for a new subject to train his/her personal authentication system. To perform this analysis, a fixed number of cycles were randomly extracted from the whole target dataset and were used to train the OSVM. The remaining walking cycles were used as the positive test set. In \fig{fig:06_OSVM_Ncyc_optimization} we show the F-measure as a function of this number of cycles. From these results, it follows that increasing the number of cycles beyond $1,000$ leads to little improvement. This number corresponds to about $15$ minutes of walking activity, distributed among different acquisition sessions. Multiple sessions are recommended to account for some statistical variation due to wearing different clothes.

\begin{figure}[t]
	\centering
	\includegraphics[width=\figw]{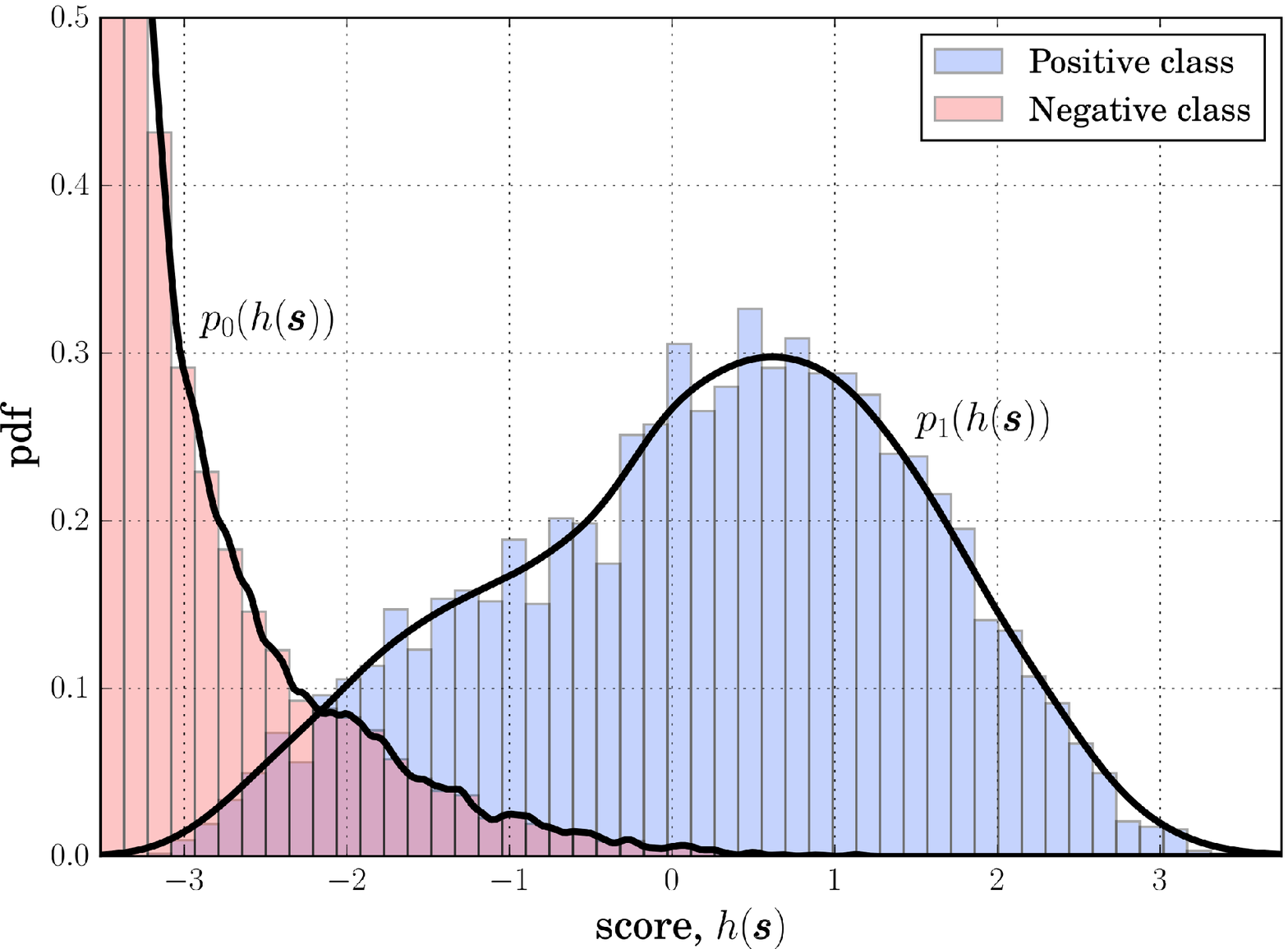}
	\caption{Empirical pdf of the OSVM scores for class $C_1$ ($p_1(\scorefunc(\pmb{s}))$) and $C_0$ ($p_0(\scorefunc(\pmb{s}))$).}
	\label{img:09_OSVM_score_hist}
\end{figure}

Once all the model's parameters are defined, the OSVM score can be analyzed. Let $p_\theta(\scorefunc(\pmb{s}))=p(\scorefunc(\pmb{s}) \ |\ \pmb{s} \in C_{\theta})$  be the estimated probability density function (pdf) of the OSVM score $\scorefunc(\pmb{s}) \in \mathbb{R}$, provided that the walking cycle belongs to a user of class $C_\theta$ with $\theta \in \{0,1\}$. Empirical pdfs $p_\theta(\scorefunc(\pmb{s}))$ from our dataset are provided in~\fig{img:09_OSVM_score_hist}.

\begin{figure*}[htpb]
	\centering
	\includegraphics[width=\textwidth]{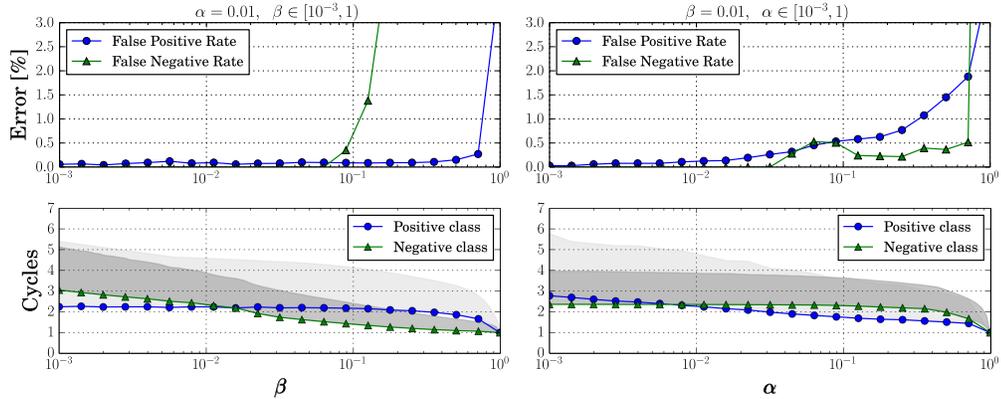}
	\caption{Results of the \mbox{multi-stage} authentication framework. False positive and negative rates are shown in the top graphs, the number of walking cycles required to make a final decision on the user's identity is shown in the bottom ones. Upper shaded areas extend for a full standard deviation from the mean and include about $80$\% of the events. }
	\label{img:10_Accumulation_results}
\end{figure*}
%


\section{Multi-Stage Authentication}
\label{sec:sequential_analysis}

The so far discussed processing pipeline returns a score for each walking cycle. However, as seen in \fig{img:09_OSVM_score_hist}, when a score falls near the point where the two pdfs intersect, there is a high uncertainty about the identity of the user who generated it. In \fraName, we resolve this indetermination by jointly considering the scores from successive walking cycles. Let $\mathcal{O} = (o_1, o_2, \dots)$ be a sequence of subsequent OSVM scores from the same subject, where $o_i = \scorefunc(\pmb{s}_i) \in \mathbb{R}$ and $i=1,2,\dots$ is the walking cycle index. From our previous analysis, $o_i$ can be thought of as a random process having probability density function $p_{\theta}(\scorefunc(\pmb{s}_i))=p_{\theta}(o_i)$, $\theta \in \{0,1\}$, and our objective is to reliably estimate $\theta$ from the scores in $\mathcal{O}$. Toward this, we assume that subsequent scores belong to the same user and that they are independent and identically distributed (i.i.d), i.e., they are independently drawn from $p_{\theta}(\cdot)$, with $\theta$ unknown.

For the estimation of $\theta$ we use Wald's probability ratio test (SPRT)~\cite{Wald-1947,Tartakovsky-2015}. We define the two hypotheses \mbox{$\{H_1 : \theta = 1\}$}, meaning that the sequence $\mathcal O$ belongs to the target user (class $C_1$), and \mbox{$\{H_0 : \theta = 0\}$}, meaning that another user generated it  (class $C_0$). Hence, we assess which one of these is true through SPRT sequential binary testing. That is, we keep measuring new scores and use them to decrease our uncertainty about $\theta$. Considering $n$ samples $(o_1,o_2,\dots,o_n)$, the final decision takes on two values $D_n = 0$ or $D_n=1$, where $D_n = j$, $j \in \{0,1\}$ means that hypothesis $H_j$ is accepted and therefore the alternative hypothesis is rejected.
Owing to our assumptions (i.i.d. scores, generated by the same subject), for $n$ scores $\mathcal{O}_n=(o_1,o_2,\dots,o_n)$ the joint pdf is:
\begin{equation}
\label{eq:joint_prob}
\tilde{p}_{\theta}(\mathcal{O}_n) = \prod_{j=1}^n p_{\theta}(o_j), \quad \theta \in \{0,1\} \, . 
\end{equation}
Defining $\lambda_j = p_1(o_j) / p_0(o_j)$, the likelihood ratio of the sequence $\mathcal{O}$ truncated at index $n$, $\mathcal{O}_n$, is 
\begin{equation}
\label{eq:LR_acc}
\frac{\tilde{p}_{1}(\mathcal{O}_n)}{\tilde{p}_{0}(\mathcal{O}_n)}  = \prod_{j=1}^n \dfrac{p_1(o_j)}{p_0(o_j)} = \prod_{j=1}^n \lambda_j \, ,
\end{equation}
and applying the logarithm, we get: 
\begin{equation}
\label{eq:LLR_acc}
\Lambda_n = \log\left( \dfrac{\tilde{p}_1(\mathcal{O}_n)}{\tilde{p}_0(\mathcal{O}_n)}  \right) = \sum_{j=1}^n \log \left( \lambda_j \right) \, .
\end{equation}
If we wait a further step $n+1$ before making a decision, from \eq{eq:LLR_acc} the new log-likelihood $\Lambda_{n+1}$ is conveniently obtained as $\Lambda_{n+1} = \Lambda_n + \log (\lambda_{n+1})$. The SPRT test starts from time \mbox{$1$}, obtaining one-class OSVM scores $o_1,o_2,\dots$ for each successive walking cycle. After $n$ cycles, the cumulative log-likelihood ratio is $\Lambda_{n} = \Lambda_{n-1} + \log (\lambda_{n})$, with $\Lambda_{0}=0$. Two suitable thresholds $A$ and $B$ are defined and the test continues to the next cycle $n+1$ if $A < \Lambda_n < B$, $H_1$ is accepted if \mbox{$\Lambda_n \geq B$}, whereas $H_0$ is accepted if \mbox{$\Lambda_n \leq A$}. Moreover, defining $\alpha$ as the probability of accepting $H_1$ when $H_0$ is true and $\beta$ that of accepting $H_0$ when $H_1$ is true, $A$ and $B$ can be approximated as: $A = \log(\beta/(1-\alpha))$ and $B=\log((1-\beta)/\alpha)$, see~\cite{Wald-1947}. 

\subsection{Experimental Results}
\label{sec:experimental_results}

The motion data from $K=35$ subjects was used to train the CNN feature extractor, with $\ncycles=40$, $F=40$ and $S=20$. One user out of the remaining $15$ was considered as the {\it target} user and $14$ as the negatives for the final tests. The following results are obtained through a leave-one-out cross-validation approach for the sessions of the target user, i.e., out of twelve sessions, eleven are used for training and one for the final tests. The session that is left out is rotated and the final results are averaged across all trials.  
The authentication results of the \mbox{multi-stage} framework are shown in Fig.~\ref{img:10_Accumulation_results}. False positive rates (i.e., a user is mistakenly authenticated as the target) and false negative ones (i.e., the target is not recognized) are smaller than $0.15$\% for an appropriate choice of the SPRT thresholds ($\alpha$ and $\beta$). Also, a reliable authentication requires fewer than five walking cycles in $80$\% of the cases. This means that the framework is very accurate and at the same time fast. We remark that the best authentication results that were obtained in previous papers lead to error rates ranging from $5$ to $15$\%~\cite{thang2012gait, nickel2012authentication, watanabe2014influence, choi2014biometric, ren2013smartphone, Sprager-15}. A comparison with these approaches is very difficult to carry out due to the different datasets (e.g., number of subjects and walking time), acquisition settings (e.g., smartphone or sensors location). The reader can nevertheless refer to Section~\ref{sec:network_optimization} for a fair comparison between our \mbox{single-step} classification framework and classical feature extraction techniques on our dataset.


As for our assumptions, in light of the small number of cycles required, it is reasonable to presume that the same subject generates the scores in $\mathcal{O}$. For the i.i.d. assumption, we extended the decision framework to the first-order autoregressive model of~\cite[Chapter 3, p. 158]{Tartakovsky-2015}, which allows tracking the correlation across successive cycles. However, this did not lead to any appreciable performance improvement and only implied a higher complexity. The reason is that scores are lightly correlated in time.


\section{Conclusions}
\label{sec:conclusions}

In this paper we have proposed \fraName, a user authentication framework for inertial signals acquired from smartphones. Various schemes performing manual feature extraction and using the selected features for user classification have appeared in the recent literature. In sharp contrast with these, \fraName\ exploits convolutional neural networks, as they allow for an automatic feature engineering and have excellent generalization capabilities. These deep neural networks are then used as universal feature extractors to feed classification techniques, combining them with \mbox{one-class} support vector machines and a novel \mbox{multi-stage} decision algorithm. With our framework, the neural network is trained once for all and subsequently utilized for new users. The \mbox{one-class} classifier is solely trained using motion data from the target subject; it returns a score weighing the dissimilarity of newly acquired data with respect to that of the target. Subsequent scores are then accumulated through a \mbox{multi-stage} decision approach.
Experimental results show the superiority of \fraName\ against prior work, leading to misclassification rates smaller than $0.15$\% in fewer than five walking cycles. Design choices and the optimization of the various processing blocks were discussed and compared against existing algorithms.


\bibliography{bibliography}

\begin{thebibliography}{10}
\expandafter\ifx\csname url\endcsname\relax
  \def\url#1{\texttt{#1}}\fi
\expandafter\ifx\csname urlprefix\endcsname\relax\def\urlprefix{URL }\fi
\expandafter\ifx\csname href\endcsname\relax
  \def\href#1#2{#2} \def\path#1{#1}\fi

\bibitem{sprager2015inertial}
S.~Sprager, M.~B. Juric, Inertial sensor-based gait recognition: A review,
  Sensors 15~(9) (2015) 22089.

\bibitem{gafurov2007survey}
D.~Gafurov, A survey of biometric gait recognition: Approaches, security and
  challenges, in: Annual Norwegian computer science conference, Oslo, Norway,
  2007.

\bibitem{zeng2014silhouette}
W.~Zeng, C.~Wang, F.~Yang, Silhouette-based gait recognition via deterministic
  learning, Pattern Recognition 47~(11) (2014) 3568--3584.

\bibitem{luo2016robust}
J.~Luo, J.~Tang, T.~Tjahjadi, X.~Xiao, {Robust arbitrary view gait recognition
  based on parametric 3D human body reconstruction and virtual posture
  synthesis}, Pattern Recognition 60 (2016) 361--377.

\bibitem{xing2016complete}
X.~Xing, K.~Wang, T.~Yan, Z.~Lv, Complete canonical correlation analysis with
  application to multi-view gait recognition, Pattern Recognition 50 (2016)
  107--117.

\bibitem{chen2016uncooperative}
X.~Chen, J.~Xu, Uncooperative gait recognition: Re-ranking based on sparse
  coding and multi-view hypergraph learning, Pattern Recognition 53 (2016)
  116--129.

\bibitem{choudhury2015robust}
S.~D. Choudhury, T.~Tjahjadi, Robust view-invariant multiscale gait
  recognition, Pattern Recognition 48~(3) (2015) 798--811.

\bibitem{whittle2007analysis}
M.~W. Whittle, Gait Analysis: An Introduction, 4th ed., Elsevier: Edinburgh,
  UK, 2008.

\bibitem{chan2012evaluating}
H.~Chan, H.~Zheng, H.~Wang, R.~Sterritt, Evaluating and overcoming the
  challenges in utilizing smart mobile phones and standalone accelerometer for
  gait analysis, in: IET Irish Signals and Systems Conference (ISSC 2012),
  Maynooth, Ireland, 2012.

\bibitem{razavian2014computer}
A.~S. Razavian, H.~Azizpour, J.~Sullivan, S.~Carlsson, {CNN Features
  Off-the-Shelf: An Astounding Baseline for Recognition}, in: {IEEE Conference
  on Computer Vision and Pattern Recognition Workshops}, Columbus, Ohio, US,
  2014.

\bibitem{Bishop-2007}
C.~Bishop, {Pattern Recognition and Machine Learning}, Springer, 2007.

\bibitem{thang2012gait}
H.~M. Thang, V.~Q. Viet, N.~D. Thuc, D.~Choi, Gait identification using
  accelerometer on mobile phone, in: International Conference on Control,
  Automation and Information Sciences (ICCAIS), Saigon, Vietnam, 2012.

\bibitem{nickel2012authentication}
C.~Nickel, T.~Wirtl, C.~Busch, Authentication of smartphone users based on the
  way they walk using k-nn algorithm, in: International Conference on
  Intelligent Information Hiding and Multimedia Signal Processing (IIH-MSP),
  Piraeus-Athens, Greece, 2012.

\bibitem{watanabe2014influence}
Y.~Watanabe, {Influence of Holding Smart Phone for Acceleration-Based Gait
  Authentication}, in: International Conference on Emerging Security
  Technologies (EST), Houston, Texas, US, 2014.

\bibitem{choi2014biometric}
S.~Choi, I.~H. Youn, R.~LeMay, S.~Burns, J.~H. Youn, Biometric gait recognition
  based on wireless acceleration sensor using k-nearest neighbor
  classification, in: International Conference on Computing, Networking and
  Communications (ICNC), Honolulu, Hawaii, US, 2014.

\bibitem{ren2013smartphone}
Y.~Ren, Y.~Chen, M.~C. Chuah, J.~Yang, Smartphone based user verification
  leveraging gait recognition for mobile healthcare systems, in: IEEE
  Communications Society Conference on Sensor, Mesh and Ad Hoc Communications
  and Networks (SECON), New Orleans, Louisiana, US, 2013.

\bibitem{Sprager-15}
S.~Sprager, M.~B. Juric, {An Efficient HOS-Based Gait Authentication of
  Accelerometer Data}, {IEEE Transactions on Information Forensics and
  Security} 10~(7) (2015) 1486--1498.

\bibitem{chan2013smart}
H.~Chan, H.~Zheng, H.~Wang, R.~Sterritt, D.~Newell, Smart mobile phone based
  gait assessment of patients with low back pain, in: Ninth International
  Conference on Natural Computation (ICNC), San Diego, California, US, 2013.

\bibitem{huang2012gait}
G.-S. Huang, C.~C. Wu, J.~Lin, Gait analysis by using tri-axial accelerometer
  of smart phones, in: International Conference on Computerized Healthcare
  (ICCH), Hong Kong, China, 2012.

\bibitem{nickel2011scenario}
C.~Nickel, M.~O. Derawi, P.~Bours, C.~Busch, Scenario test of
  accelerometer-based biometric gait recognition, in: International Workshop on
  Security and Communication Networks (IWSCN), Gj\o{}vik, Norway, 2011.

\bibitem{nickel2011using}
C.~Nickel, C.~Busch, S.~Rangarajan, M.~Mobius, Using hidden markov models for
  accelerometer-based biometric gait recognition, in: IEEE 7th International
  Colloquium on Signal Processing and its Applications (CSPA), Penang,
  Malaysia, 2011.

\bibitem{kobayashi2011rotation}
T.~Kobayashi, K.~Hasida, N.~Otsu, {Rotation invariant feature extraction from
  3-D acceleration signals}, in: IEEE International Conference on Acoustics,
  Speech and Signal Processing (ICASSP), Prague, Czech Republic, 2011.

\bibitem{scholkopf2001SVM}
B.~Sch\"{o}lkopf, J.~C. Platt, J.~C. Shawe-Taylor, A.~J. Smola, R.~C.
  Williamson, Estimating the support of a high-dimensional distribution, Neural
  Computation 13~(7) (2001) 1443--1471.

\bibitem{murray1964walking}
M.~P. Murray, A.~B. Drought, R.~C. Kory, Walking patterns of normal men, The
  Journal of Bone \& Joint Surgery 46~(2) (1964) 335--360.

\bibitem{murray1967gait}
M.~P. Murray, Gait as a total pattern of movement: Including a bibliography on
  gait., American Journal of Physical Medicine \& Rehabilitation 46~(1) (1967)
  290--333.

\bibitem{nixon2006human}
T.~Nixon, M. S. ans~Tieniu, C.~Rama, Human identification based on gait,
  Springer, 2006.

\bibitem{kwapisz2010cell}
J.~R. Kwapisz, G.~M. Weiss, S.~A. Moore, Cell phone-based biometric
  identification, in: Fourth IEEE International Conference on Biometrics:
  Theory Applications and Systems (BTAS), 2010.

\bibitem{mantyjarvi2005identifying}
J.~Mantyjarvi, M.~Lindholm, E.~Vildjiounaite, S.~M. Makela, H.~A. Ailisto,
  Identifying users of portable devices from gait pattern with accelerometers,
  in: IEEE International Conference on Acoustics, Speech, and Signal Processing
  (ICASSP), Philadelphia, Pennsylvania, US, 2005.

\bibitem{derawi2010unobtrusive}
M.~O. Derawi, C.~Nickel, P.~Bours, C.~Busch, Unobtrusive user-authentication on
  mobile phones using biometric gait recognition, in: 6th International
  Conference on Intelligent Information Hiding and Multimedia Signal Processing
  (IIH-MSP), Darmstadt, Germany, 2010.

\bibitem{DTW-2005}
E.~Keogh, C.~Ratanamahatana, Exact indexing of dynamic time warping, {Knowledge
  and Information Systems} 7~(3) (2005) 358--386.

\bibitem{juefei2012gait}
F.~Juefei-Xu, C.~Bhagavatula, A.~Jaech, U.~Prasad, M.~Savvides, Gait-id on the
  move: Pace independent human identification using cell phone accelerometer
  dynamics, in: Fifth International Conference on Biometrics: Theory,
  Applications and Systems (BTAS), Washington DC, US, 2012.

\bibitem{jiang2013possibility}
S.~Jiang, B.~Zhang, G.~Zou, D.~Wei, The possibility of normal gait analysis
  based on a smart phone for healthcare, in: IEEE International Conference on
  Green Computing and Communications (GreenCom), Internet of Things (iThings),
  and Cyber, Physical and Social Computing (CPSCom), Beijing, China, 2013.

\bibitem{zhong2014sensor}
Y.~Zhong, Y.~Deng, Sensor orientation invariant mobile gait biometrics, in:
  IEEE International Joint Conference on Biometrics (IJCB), Clearwater, FL,
  USA, 2014.

\bibitem{CNN_video_analysis_2014}
A.~Karpathy, G.~Toderici, S.~Shetty, T.~Leung, R.~Sukthankar, L.~Fei-Fei,
  {Large-scale Video Classification with Convolutional Neural Networks}, in:
  {IEEE Conference on Computer Vision and Pattern Recognition (CVPR)},
  {Columbus, Ohio, US}, 2014.

\bibitem{ngo2014largest}
T.~T. Ngo, Y.~Makihara, H.~Nagahara, Y.~Mukaigawa, Y.~Yagi, The largest
  inertial sensor-based gait database and performance evaluation of gait-based
  personal authentication, Pattern Recognition 47~(1) (2014) 228--237.

\bibitem{casale2012personalization}
P.~Casale, O.~Pujol, P.~Radeva, Personalization and user verification in
  wearable systems using biometric walking patterns, Personal and Ubiquitous
  Computing 16~(5) (2012) 563--580.

\bibitem{tilmanne2008adatabase}
J.~Tilmanne, R.~Sebbe, T.~Dutoit, A database for stylistic human gait modeling
  and synthesis (2008).

\bibitem{frank2010data}
J.~Frank, S.~Mannor, D.~Precup,
  \href{http://www.cs.mcgill.ca/~jfrank8/data/gait-dataset.html}{Data sets:
  Mobile phone gait recognition data} (2010).
\newline\urlprefix\url{http://www.cs.mcgill.ca/~jfrank8/data/gait-dataset.html}

\bibitem{Welch-67}
P.~D. Welch, The use of fast fourier transform for the estimation of power
  spectra: A method based on time averaging over short, modified periodograms,
  {IEEE Transactions on Audio and Electroacoustics} 15~(2) (1967) 70--73.

\bibitem{teixeira2009distributed}
T.~Teixeira, D.~Jung, G.~Dublon, A.~Savvides, Pem-id: Identifying people by
  gait-matching using cameras and wearable accelerometers, in: ACM/IEEE
  International Conference on Distributed Smart Cameras (ICDSC), Como, Italy,
  2009.

\bibitem{pca-heading-2009}
K.~Kunze, P.~Lukowicz, K.~Partridge, B.~Begole, {Which Way Am I Facing:
  Inferring Horizontal Device Orientation from an Accelerometer Signal}, in:
  {IEEE International Symposium on Wearable Computers}, {Linz, Austria}, 2009.

\bibitem{heading-2015}
Z.-A. Deng, G.~Wang, Y.~Hu, D.~Wu, {Heading Estimation for Indoor Pedestrian
  Navigation Using a Smartphone in the Pocket}, {MDPI Sensors} 15~(9) (2015)
  21518--21536.

\bibitem{PCA-64}
C.~R. Rao, {The Use and Interpretation of Principal Component Analysis in
  Applied Research}, {Sankhy\={a}: The Indian Journal of Statistics} 26~(4)
  (1964) 329--358.

\bibitem{lecun1998convolutional}
Y.~LeCun, Y.~Bengio, Convolutional networks for images, speech, and time
  series, in: The Handbook of Brain Theory and Neural Networks, MIT Press,
  1998, pp. 255--258.

\bibitem{krizhevsky2012imagenet}
A.~Krizhevsky, I.~Sutskever, G.~E. Hinton, Imagenet classification with deep
  convolutional neural networks, in: Advances in Neural Information Processing
  Systems 25, 2012, pp. 1106--1114.

\bibitem{scherer2010evaluation}
D.~Scherer, A.~M\"{u}ller, S.~Behnke, Evaluation of pooling operations in
  convolutional architectures for object recognition, in: 20th International
  Conference on Artificial Neural Networks (ICANN), Thessaloniki, Greece, 2010.

\bibitem{hanka1997curse}
R.~Hanka, T.~P. Harte, Computer Intensive Methods in Control and Signal
  Processing: The Curse of Dimensionality, Birkh{\"a}user Boston, 1997, Ch.
  Curse of Dimensionality: Classifying Large Multi-Dimensional Images with
  Neural Networks, pp. 249--260.

\bibitem{quinlan1993c45}
J.~R. Quinlan, C4.5: Programs for Machine Learning, Morgan Kaufmann Publishers
  Inc., San Francisco, California, US, 1993.

\bibitem{friedman1997bayesian}
N.~Friedman, D.~Geiger, M.~Goldszmidt, Bayesian network classifiers, Machine
  Learning 29~(2) (1997) 131--163.

\bibitem{cover1967nearest}
T.~Cover, P.~Hart, Nearest neighbor pattern classification, IEEE Transactions
  on Information Theory 13~(1) (1967) 21--27.

\bibitem{cortes1995support}
C.~Cortes, V.~Vapnik, {Support-vector networks}, {Machine Learning} 20~(3)
  (1995) 273--297.

\bibitem{scholkopf1999support}
B.~Sch{\"o}lkopf, R.~C. Williamson, A.~J. Smola, J.~Shawe-Taylor, J.~C. Platt,
  et~al., Support vector method for novelty detection, Neural Information
  Processing Systems (NIPS) 12 (1999) 582--588.

\bibitem{Fmeasure-2003}
D.~R. Musicant, V.~Kumar, A.~Ozgur, {Optimizing F-Measure with Support Vector
  Machines}, in: {16-th International FLAIRS Conference}, FLAIRS, {St.
  Augustine, Florida, US}, 2003.

\bibitem{tax2003artificial}
D.~M.~J. Tax, K.~R. M{\"u}ller, Artificial Neural Networks and Neural
  Information Processing, Springer, Berlin, Heidelberg, 2003, Ch. Feature
  Extraction for One-Class Classification, pp. 342--349.

\bibitem{Wald-1947}
A.~Wald, Sequential analysis, Dover, {New York, NY, US}, 1947.

\bibitem{Tartakovsky-2015}
A.~Tartakovsky, I.~Nikiforov, M.~Basseville, {Sequential Analysis Hypothesis
  Testing and Changepoint Detection}, CRC Press, 2015.

\end{thebibliography}

\end{document}